\documentclass[12pt]{article}      

\advance\voffset by -0.25cm
\usepackage{graphicx,amsmath,amsfonts,amsthm,bbm,setspace,url,verbatim,mathrsfs,xfrac,enumitem}
\usepackage{mymacros}
\usepackage{subcaption,xcolor}
\usepackage{hyperref,cleveref}
\usepackage{authblk}
\usepackage[english]{babel}
\usepackage{natbib}

\title{\textbf{\Large Fast Likelihood-Free Parameter Estimation for L\'evy Processes} \thanks{Acknowledgment: This research was supported by National Science Foundation grant DMS-2310487.}}

\author[1]{Nicolas Coloma\thanks{Email: Nicolas.ColomaCarphio@colorado.edu }}
\author[1]{William Kleiber\thanks{Email: William.Kleiber@colorado.edu}}

\affil[1]{\small Department of Applied Mathematics, University of Colorado, Boulder, CO, USA}

\date{\small \today} 

\begin{document}
\doublespacing
\maketitle

\begin{abstract}
\small
L\'evy processes are widely used in financial modeling due to their ability to capture discontinuities and heavy tails, which are common in high-frequency asset return data.  
However, parameter estimation remains a challenge when associated likelihoods are unavailable or costly to compute. 
We propose a fast and accurate method for L\'evy parameter estimation using the neural Bayes estimation (NBE) framework -- a simulation-based, likelihood-free approach that leverages permutation-invariant neural networks to approximate Bayes estimators. 
We contribute new theoretical results, showing that NBE results in consistent estimators whose risk converges to the Bayes estimator under mild conditions.
Moreover, through extensive simulations across several L\'evy models, we show that NBE outperforms traditional methods in both accuracy and runtime, while also enabling two complementary approaches to uncertainty quantification. 
We illustrate our approach on a challenging high-frequency cryptocurrency return dataset, where the method captures evolving parameter dynamics and delivers reliable and interpretable inference at a fraction of the computational cost of traditional methods. 
NBE provides a scalable and practical solution for inference in complex financial models, enabling parameter estimation and uncertainty quantification over an entire year of data in just seconds. 
We additionally investigate nearly a decade of high-frequency Bitcoin returns, requiring less than one minute to estimate parameters under the proposed approach.

\end{abstract}

\section{Introduction}

Modeling and analysis of financial time series has become increasingly important, particularly with the rise of complex, high-frequency datasets that often exhibit jumps and heavy-tailed behavior. 
These issues are especially pronounced in emerging markets like cryptocurrencies, which are traded continuously across global exchanges and exhibit frequent fluctuations unconstrained by traditional market hours.

The growing interest in cryptocurrencies has spurred a surge of modeling approaches aimed at capturing their dynamics. 
For instance, \citet{cocco2017} developed a generative model for Bitcoin prices, while \citet{chu2020} examined momentum trading strategies using hourly data. 
\citet{han2022} introduced a simulation model for asset prices, and \citet{celeste2020} explored wavelet-based models to study the volatility of Ethereum and Bitcoin returns.

This recent work builds on a broader literature focused on modeling high-frequency and heavy-tailed financial data. 
Additional contributions include the assessment of the distributional properties of daily returns \citep{andersen2010} and the development of dynamic discrete copula models for high-frequency stock price changes \citep{koopman2018}. 
These advances highlight the need for flexible modeling frameworks capable of capturing discontinuities, jump behavior, and volatility features—characteristics that are especially prominent in modern cryptocurrency markets.

To better accommodate complex behavior such as discontinuities and heavy tails, financial researchers have long turned to L\'evy processes -- a flexible class of stochastic models characterized by stationary and independent increments. 
These processes are well established in financial econometrics and have been foundational for modeling returns for decades \citep{bertoin1996, madan1998,sato1999, contTankov2004, applebaum2009}.

Despite their clear utility for cryptocurrency modeling, L\'evy processes have only recently been explored in this context. 
For instance, \citet{shirvani2024} proposed a doubly subordinated normal inverse Gaussian L\'evy process for daily Bitcoin return data, while \citet{dvgBerry2023} demonstrated that deeper subordination improves both tail modeling and small log-return behavior on high-frequency data. 

Although such models offer compelling fits, fast and scalable parameter estimation remains a persistent challenge \citep{ sueishi2005, kunitomo2006, belomestny2015}, primarily because many L\'evy models lack tractable likelihoods. 
This is especially true for more expressive models, such as subordinated processes \citep{dvgBerry2023, shirvani2024}, which have been shown to capture both central and tail behavior effectively in many financial datasets, but do not admit closed-form densities.
As a result, classical likelihood-based methods are either computationally expensive or simply not applicable.

To address this, a range of alternative methods have been developed, including those based on the empirical characteristic function, such as minimum distance estimators \citep{press1972} and generalized method of moments (GMM) and its extensions \citep{feuerverger1981, carrasco2000}. 
Composite likelihood techniques provide another workaround by combining marginal or conditional likelihoods \citep{cox2004, varin2005, varin2011}. 
Additional approaches include maximum empirical likelihood \citep{elgin2011} and nonparametric techniques \citep{figueroa2009}.
While these methods are valuable, they often suffer from reduced statistical efficiency \citep{castruccio2016} and tend to be computationally intensive, especially in high-frequency applications.

Given these limitations, simulation-based Bayesian approaches such as approximate Bayesian computation (ABC) have become an attractive alternative \citep{chon1997,beaumont2002, sisson2018}. 
Recent developments in ABC have integrated neural networks to improve flexibility and scalability. 
In particular, there has been substantial work on neural point estimators: neural networks designed to learn a direct mapping from the observed data to the parameter space, producing point estimates rather than full posterior distributions. 
Such neural methods have been successfully applied across various domains, including genetics, ecology, econometrics, and spatio-temporal modeling \citep{creel2017neural, flagel2018, zammit2020, gerber2021, banesh2021, rudi2022, lenzi2023}. 
These developments suggest that neural-based methods could offer a scalable alternative for L\'evy parameter estimation.

Neural networks have also gained significant attention within the financial community, where they are valued for the improvements in speed and accuracy they can provide. 
Substantial work has focused on their application to model calibration and option pricing.
For instance, \cite{hernandez2016} demonstrated that neural networks can be used effectively for model calibration in settings where the likelihood is intractable. 
Subsequent works, such as \cite{liu2019, gan2020}, further developed neural frameworks for model calibration. 
\cite{horvath2021} introduced a method in which neural networks approximate complex pricing functions that are otherwise difficult to evaluate.
See \citep{ruf2019} for a review of this growing literature on neural networks for option pricing and hedging.  

More recently, neural networks have also been applied directly in the context of L\'evy processes within the financial literature. 
For example, \cite{huh2019} proposed a new non-parametric model, the exponential L\'evy neural network (ELNN), for option pricing.
\cite{kudryavtsev2020,kudryavtsev2023} employed neural networks to simulate and calibrate different classes of L\'evy processes. 
These works underscore a growing trend at the intersection of machine learning and L\'evy process modeling. 

In this work, we propose to use the neural Bayes estimation (NBE) framework \citep{sainsbury2023nbe}, a simulation-based approach that leverages neural networks to approximate Bayes estimators without requiring access to a likelihood. 
Our contributions are fourfold.
First, we introduce a flexible neural Bayes estimator for L\'evy process parameters that bypasses the need for likelihood evaluation.
Second, we implement the estimator using a DeepSets-based architecture \citep{zaheer2017deepSets}, which is tailored for inference from replicates of unordered i.i.d. data.
Third, we provide new theoretical results showing that the estimator is consistent and converges in risk to the Bayes estimator, offering formal guarantees on its large-sample behavior.
Fourth, we extend uncertainty quantification by comparing two complementary approaches — bootstrap-based confidence intervals and posterior-quantile estimation — thereby providing a principled and interpretable framework for likelihood-free inference.

We evaluate this approach through several simulation studies involving common L\'evy models and demonstrate that it consistently outperforms benchmark methods in both accuracy and computational efficiency.
Finally, we apply the method to high-frequency cryptocurrency return data, where NBE delivers fast and accurate parameter estimation, outperforming traditional methods in both runtime and statistical precision. 
Remarkably, our approach allows us to estimate parameters and construct confidence intervals for an entire year of high-frequency data in just seconds, marking a significant computational improvement over existing techniques.

\section{Framework}

\subsection{Background on L\'evy Processes}

L\'evy processes are a class of stochastic processes characterized by stationary and independent increments.
Formally, a stochastic process $Z(t),t\geq0$, is a L\'evy process if it satisfies the following conditions:
\begin{itemize}
   \item $Z(0)=0$ almost surely.
    \item For any $0\leq t_1 < t_2 < \ldots < t_n$, the random variables $Z(t_1)$, $Z(t_2)-Z(t_1)$ , $\ldots$ , $Z(t_n)-Z(t_{n-1})$ are independent.
    \item The distribution of $Z(t+h)-Z(t)$ does not depend on $t$.
    \item $Z$ is stochastically continuous.
\end{itemize}
L\'evy processes are closely connected to the class of infinitely divisible distributions \citep{sato1999}. 
Moreover, the L\'evy-Khintchine theorem provides a unique characterization of the characteristic function of a L\'evy process.
\begin{theorem}[L\'evy-Khintchine]\label{theo:Levy}
    \small
        Let $Z(t)$ be a L\'evy process on $\mathbb{R}$, then 
        \begin{align} \nonumber
        \mathbb{E}[\exp(\mathrm{i}\omega Z(t))] &= \exp \Bigg( -\frac{t}{2} {\sigma}^2\omega^2 + \mathrm{i} t{\gamma}\omega  \nonumber \\
        &\qquad + t \int \left[e^{\mathrm{i}\omega x}-1-\mathrm{i}\omega x  \mathbbm{1}_{[|x|\leq1]}(x) \right] {\nu}(dx) \Bigg) \nonumber \\
        &=\exp( t \Psi(w)),
        \end{align}
        for $\omega \in \mathbb{R}$, where $\gamma \in \mathbb{R}$, $\sigma \geq 0$, and $\nu$ is a measure on $\mathbb{R}$ such that $\nu(\{0\})=0$ and $\int ( |x|^2 \wedge 1)\nu(dx)  < \infty$ . 
\end{theorem}
In this theorem, $\mathbbm{1}$ denotes the indicator function, $\nu$ is the L\'evy measure that governs the jump behavior of the process, $\gamma$ and $\sigma^2$ represent the drift and variance of the process, respectively, and $\Psi$ is the characteristic exponent. 
The triplet $\left(\gamma, \sigma, \nu\right)$ is referred to as the characteristic triplet of the L\'evy process $Z(t)$ \citep{contTankov2004}, and fully defines the process structure.

L\'evy processes have been widely used to model financial assests. 
Denote the price of an asset by $S_t$. 
This price is then often represented by $S_t=S_0\exp{X_t}$ where $X_t$ is a L\'evy process. 
Consequently, the log returns are given by $\log\left( \frac{S_t}{S_{t-1}} \right)=X_t-X_{t-1}$. 

\subsubsection{Examples of Common Models}\label{sec:simplerModels}

For a comprehensive discussion of the properties of L\'evy processes and infinitely divisible distributions, see \citep{sato1999, contTankov2004}.
Despite years of development, new classes of L\'evy processes continue to be explored.
In this section, we focus on three widely studied examples: the compound Poisson process, the Merton model, and the variance gamma process.  

The compound Poisson process is a pure jump process, where jumps occur randomly at times according to a Poisson process.  
Formally, it can be written as
\[
  Z(t) = \sum_{i=1}^{N(t)} Y_i,
\]
where $N(t)$ is a Poisson process with intensity $\lambda$, and $\{ Y_i \}$ are independent and identically distributed jump sizes drawn from a specified distribution, in our case, $Y_i \sim N(\mu,\sigma^2)$. 
This is an example of an infinite activity model \citep{contTankov2004}, in which the process moves essentially by jumps.

The second model is termed the Merton model \citep{merton1976}.
This model can be thought of extending the compound Poisson process by incorporating continuous fluctuations.  
It is given by
\[
  Z(t) = \mu t + \sigma W(t) + \sum_{i=1}^{N(t)} Y_i,
\]
where $W(t)$ is standard Brownian motion, $\mu$ is the drift, $\sigma$ the volatility, $N(t)$ a Poisson process with intensity $\lambda$, and $Y_i$ are normally distributed jump sizes with mean $\mu_J$ and variance $\sigma_J^2$.  
This model accommodates small Gaussian fluctuations and rare, large jumps, making it particularly attractive for asset returns.

Finally, the variance gamma (VG) process \citep{madan1998} is a time-changed Brownian motion, where time randomly expands or contracts according to a gamma process.  
Specifically,
\[
Z(t) = \gamma G(t) + \sigma W(G(t)),
\]
where $W(t)$ is standard Brownian motion, $G(t)$ is a gamma process with mean rate $\alpha$ and variance rate $\alpha$, and $\gamma$ and $\sigma$ are the drift and volatility parameters of the Brownian motion.
The VG process produces returns with heavier tails and higher kurtosis than Gaussian models, while maintaining finite variance, making it a popular model for financial time series.

These three models serve as useful benchmarks for evaluating the performance of the proposed estimation methodology under different types of jump behavior and volatility structures.

\subsubsection{Deep Variance Gamma Processes} \label{sec:dvg}

A notable subclass of L\'evy processes is that of subordinators: non-negative, almost surely increasing processes in time \citep{contTankov2004}. 
The key property of subordinators is that if $Z(t)$ is a L\'evy process and $S(t)$ is a subordinator, then the time-changed process $Z(S(t))$ is also a L\'evy process. 
This property allows subordinators to serve as building blocks for constructing more flexible L\'evy processes via random time changes (see, for example, the VG process defined in Section \ref{sec:simplerModels}). 

In financial modeling, these time changes are often interpreted as market time, reflecting the notion that market activity does not evolve uniformly in calendar time. 
During periods of intense trading or volatility, market time accelerates; in calmer periods, it slows down. 
This dynamic is particularly relevant for high-frequency data settings, where price movements occur unevenly over time.

An important example of a subordinated L\'evy process is the deep variance gamma (DVG) process, introduced in \citep{dvgBerry2023}.
Let $Z(t)$ denote a Brownian motion with zero drift and variance $\sigma^2$. 
Let  $S_1(t),\ldots, S_L(t)$ be $L$ independent gamma subordinators, where each $S_k(1)$ has parameter $\alpha_k$ for $k = 1,\ldots, L$. 
The DVG process of order $L$ is then defined as $Y(t) = Z(S_1(S_2(\ldots(S_L(t)))))$. 
This multiply-subordinated process offers significant modeling flexibility, enabling the process to capture heavy-tailed behavior and sharp peaks around zero—characteristics commonly observed in financial data \citep{carr2004}.

As shown in \citet{dvgBerry2023}, the DVG model provides a strong framework for capturing the empirical behavior of high-frequency cryptocurrency datasets. 
However, for orders $L \geq 2$, the model lacks a closed-form likelihood expression, which presents challenges for standard likelihood-based inference methods.

\subsection{Estimation Methods} \label{sec:estMethods}

We now turn to the problem of estimating parameters from discretely-observed L\'evy processes. 
This topic has been widely studied, with most traditional methods relying on the likelihood function \citep{barndorff1981,contTankov2004, sueishi2005}. 
However, for many L\'evy models, including the DVG process with depth $L \geq 2$, the likelihood function cannot be expressed in closed form \citep{diggle1984, dvgBerry2023}. 
In these situations, alternative approaches such as maximum empirical likelihood \citep{elgin2011} and nonparametric estimation techniques \citep{figueroa2009} have been proposed. 
We focus on two such classical approaches, namely the least squares characteristic function (LSQ) and maximum empirical likelihood (MELE) methods.
In addition, we introduce Neural Bayes Estimators (NBEs) and describe how they can be applied to parameter estimation for L\'evy processes.

\subsubsection{Classical Methods} \label{sec:classicalMethods}

Assume we observe data $\mathcal{X} = \{X_j\}_{j=1}^n$ generated from a L\'evy process, we begin with the LSQ characteristic function approach. 
This method estimates parameters by minimizing the squared average difference between the empirical and theoretical characteristic functions across a selected set of frequencies \citep{yu2004, kappus2010, xu2020}. 
This method was applied successfully to the DVG process in \citep{dvgBerry2023}, where it showed promising results in terms of accuracy.

Let $\phi(\omega,\theta)$ denote the theoretical characteristic function of a L\'evy process with parameter vector $\theta$, evaluated at frequency $\omega \in \mathbb{R}$. 
The empirical characteristic function, $\phi_n(\omega)$, is computed from the dataset as $\phi_n(\omega)=\frac{1}{n-1}\sum_{j=1}^{n-1} \text{exp}(i\omega \widetilde{X_j}) $.
Where $\widetilde{X_j}=X_{j+1}-X_j$ for $j=1,\ldots,n-1$.
The LSQ method then estimates $\hat{\theta}$ by minimizing the discrepancy between the empirical and theoretical characteristic functions,
\[
  \mathcal{L} \left(\theta \right)= \sum_{k=1}^{K} |\phi(\omega_k,\theta) -\phi_n(\omega_k)|^2
\]
where $\{\omega_k\}_{k=1}^{K}$ is a chosen set of frequencies. 
This loss function captures how well the model matches the observed data in the frequency domain. 

A second widely used method is MELE \citep{qin1994}, which also makes use of the empirical characteristic function. 
The core idea behind MELE is to assign a probability $p_j$ to each observed data point $\widetilde{X_j}$, and then find the set of probabilities that best matches a set of moment conditions derived from the characteristic function. 

Formally, MELE solves the following constrained optimization problem:
\[
  \max_{p_1,\dots,p_n} \; \sum_{i=1}^n \log p_i
\]
subject to
\[
  p_i \geq 0, 
  \quad \sum_{i=1}^n p_i = 1, 
  \quad \sum_{i=1}^n p_i \, g(X_i,\theta) = 0,
\]
where $g(X_i,\theta)$ represent the moment conditions derived from the model. 
In the L\'evy setting, these conditions are given by 
\[
  g(X_i,\theta) = e^{i\omega X_i} - \phi(\omega,\theta).
\]
In other words, MELE chooses the probabilities $p_i$ so that the weighted empirical characteristic function matches the theoretical characteristic function implied by $\theta$. 

Both LSQ and MELE require selecting a suitable set of frequencies and rely on empirical approximations to estimate the underlying parameters. 
However, a key limitation of these approaches -- common to many likelihood-free methods -- is their computational burden. 
The optimization procedures involved can be time-consuming, especially when estimating parameters repeatedly across many datasets or over a large parameter space.
This computational cost poses challenges for applications in high-frequency finance or real-time analysis, where rapid inference is essential. 
To address this challenge, more efficient estimation techniques must be considered, which we introduce in the next section.

\subsubsection{Neural Bayes Estimation} \label{sec:NBE}

One particularly attractive property of L\'evy processes is that they are often straightforward to simulate, lending themselves well to simulation-based estimation techniques. 
Neural networks have recently emerged as a powerful tool for likelihood-free inference. 
\citet{sainsbury2023nbe}  demonstrated the effectiveness of neural point estimators for parameter estimation from replicated samples, using different spatial models such as Gaussian processes, max-stable processes, and conditional extremes models.

For a given estimator $\hat{\theta}$ and dataset $\mathcal{X}$ drawn from a distribution with density $f \left(x|\theta \right)$, we consider a nonnegative loss function, $L\left( \theta, \hat{\theta}(\mathcal{X}) \right)$, which quantifies the estimation error. 
Bayes estimators aim to minimize the Bayes risk, defined as
\begin{equation} \label{eq:BayesRisk}
R_{\Omega}\left( \hat{\theta}(\cdot) \right) = \int_{\Theta} \left(\int_{\mathcal{S}} L\left( \theta, \hat{\theta}(\mathcal{X}) \right)f \left(x|\theta \right) dx \right) d\Omega(\theta)    
\end{equation}
where $\Omega(\cdot)$ is a prior distribution for $\theta$, and $\mathcal{S}$ is the set of all possible data realizations. 
In our setting, since the data is real-valued, this sample space is $\mathcal{S}=\mathbb{R}^n$.

However, Bayes estimators are rarely available in closed form for complex models. 
This is where neural networks become useful.
Neural Bayes estimation (NBE) aims to construct a neural point estimator $\hat{\theta}(\mathcal{X}, \zeta)$, where $\zeta$ contains the neural-network parameters, that approximates Bayes estimators. 
The objective is to train the network by solving the optimization problem
\[ 
  \zeta^{*} \equiv  \arg\,\min\limits_{\zeta} \   R_{\Omega}\left( \hat{\theta}(\cdot, \zeta) \right) .
\]
Since the Bayes risk is generally intractable, it can be approximated using Monte Carlo methods. 
Specifically, given a set $\vartheta$ of $K$ parameter samples from the prior $\Omega(\cdot)$ and $J$ datasets from $f \left(x|\theta \right)$ for each $\theta \in \vartheta$, collected in the set  $\mathcal{X}_\theta$, we have 
\begin{equation} \label{eq:MCApprox}
     R_{\Omega}\left( \hat{\theta}(\cdot, \zeta)\right)  \approx \frac{1}{K} \sum_{\theta \in \vartheta} \frac{1}{J}\sum_{\mathcal{X}\in \mathcal{X}_\theta} L\left( \theta, \hat{\theta}(\mathcal{X}, \zeta) \right). 
\end{equation} 

Note that this approach does not require evaluation or explicit knowledge of the likelihood function.
After training, we refer to the fitted neural point estimator as a neural Bayes estimator. 
Although training the neural network can be computationally expensive, the cost is amortized over time -- that is, there is an upfront cost in training the network, but executing the fitted network on new datasets is very fast. 
Once trained, the model can generate parameter estimates for new datasets with minimal computational overhead while also providing fast uncertainty quantification.

\subsubsection{Loss Functions} \label{sec:lossFct}

It is well known that in Bayesian decision theory, the choice of a loss function determines which functional of the posterior distribution is approximated by the Bayes estimator \cite{lehmann1998}. 
For example, under squared error loss, $L(\theta,\hat\theta)=\| \theta-\hat\theta \|^2$, a simple derivation shows that the risk is minimized when the estimator is the posterior mean. 
Thus, the squared error loss is said to be paired with the posterior mean. 

This observation is particularly relevant in the context of neural Bayes estimators, since the loss function chosen during training directly determines which posterior functional the network learns to approximate. 
For example, using the asymmetric lin-lin loss with different $\alpha$ values allows NBE to approximate posterior $\alpha$-quantiles, and thereby construct likelihood-free credible intervals. 

\subsubsection{Uncertainty Quantification} \label{sec:UQ}

Once the neural Bayes estimator is trained, uncertainty quantification (UQ) can be performed in multiple ways. 
We consider two complementary approaches below. 

Following \citet{sainsbury2023nbe}, uncertainty can be quantified within the NBE framework using a nonparametric bootstrap procedure.
In this approach, multiple bootstrap datasets are generated by resampling (with replacement) from the observed data, and the neural Bayes estimator is reevaluated for each dataset.
Confidence intervals can then be obtained from the empirical distribution of these estimates. 
Such intervals represent uncertainty in the NBE procedure itself: if $\hat{\theta}(\mathcal{X})$ denotes the estimator trained on data $\mathcal{X}$, bootstrap-based variance (for example) approximates
$\mathrm{Var}\big(\hat{\theta}(\mathcal{X})\big)$
capturing the sampling variability of the estimator, as opposed to other aspects of the posterior distribution.

Alternatively, uncertainty can be quantified by directly interrogating aspects of the posterior distribution, leveraging the connection between loss functions and Bayes estimators.  
For example, training the network with the lin-lin loss 
\[
  L(\theta, a) = \big(I\{\theta \leq a\} - \alpha\big)(a - \theta), \quad \alpha \in (0,1),
\]
yields the $\alpha$-quantile of the posterior distribution. 
By training separate networks for different values of $\alpha$, one can recover posterior credible intervals that reflect the Bayesian parameter uncertainty captured by the posterior itself. 

A trade-off between these approaches lies in computational cost. 
Estimating multiple posterior quantiles can be computationally intensive, whereas bootstrap-based UQ is generally faster, especially when simulations are readily available. 
However, bootstrap only captures the variability of the estimator itself, rather than the full posterior uncertainty. 
For a more detailed discussion on estimating the posterior distribution, or guidance on when each method may be preferred, we refer the reader to \citet{zammit2025}, and \citet{sainsbury2025}.

\subsubsection{DeepSets Architecture for L\'evy Processes} \label{sec:deepSets}

Having defined the NBE framework and the role of the loss function, we now describe the specific neural network architecture used in this work. 
Here, architecture refers to the actual structure of the neural network that determines how inputs are mapped to outputs.

A unique aspect of L\'evy processes that we wish to exploit is their stationary and independent increments. 
That is, the distribution of the vector $\widetilde{\mathbf{X}}= (\widetilde{X}_1,\widetilde{X}_2,\ldots,\widetilde{X}_{n-1})\T = (X_2-X_1,X_3-X_2,\ldots,X_n-X_{n-1})\T$ is invariant to permutations. 
Thus it is desirable to use a neural network architecture that exploits this invariance. 
We follow \citet{sainsbury2023nbe}, which is based on the DeepSets framework \citep{zaheer2017deepSets}. 
DeepSets enforces the permutation invariance property by structuring the neural estimator as follows:
\begin{equation}
  \begin{aligned}
    & \hat{\theta}(\mathcal{X}, \zeta) = \phi \left(T\left(\mathcal{X}; \zeta_{\Lambda}\right) ; \zeta_{\phi} \right) , \\
    &T\left(\mathcal{X}; \zeta_{\Lambda}\right) = a\left( \left\{ \Lambda\left(\widetilde{X}_i; \zeta_{\Lambda}\right)   \right\}_{i=1}^{n-1} \right) ,
  \end{aligned}
\end{equation}
with $\phi$ and $\Lambda$ being (deep) neural networks parameterized by $\zeta_{\phi}$ and $\zeta_{\Lambda}$, respectively, and $a(\cdot)$ a permutation-invariant function, typically a sum or average.

In this framework, each data point $\widetilde{X}_i$ is first processed through a neural network known as the summary network ($\Lambda$), which maps the raw data into a learned feature space. 
These transformed features are then aggregated using a permutation-invariant function $a(\cdot)$. 
Some examples of permutation-invariant functions include the arithmetic mean, quantiles, product and minimum or maximum. 
The aggregated result is then passed through a second neural network, the inference network ($\phi$), which produces the final parameter estimates. 
It should be noted that, under this setup, the network is set to accommodate input dimension of size $n$, and would need to be retrained for other dimensions. 
\cite{sainsbury2023nbe} suggest some adaptations to accommodate different input dimensions, but we focus on a fixed input size for the remainder of the paper. 

\subsection{Theoretical Guarantees for DeepSets-based Bayes Estimation} \label{sec:proofNBE}

In this section, we state a new consistency result showing that a sufficiently expressive DeepSets class trained by minimizing a Monte Carlo approximation to the Bayes risk recovers said risk. 

Let $\Theta\subset\mathbb{R}^d$ be the parameter space and let $\mathcal{X}=(X_1,\dots,X_n)$ be i.i.d.\ replicates with distribution determined by $\theta\in\Theta$.
Let $f(x\mid\theta)$ denote the data-generating density for $\mathcal{X}$ given $\theta$.
Recall that a Bayes estimator (for the chosen loss) is any minimizer of the Bayes risk (\ref{eq:BayesRisk}), 
\[
\hat{\theta}^\star \in \arg\min_{\hat{\theta}}  R_{\Omega}\left( \hat{\theta}(\cdot) \right).
\]

\noindent For notational simplicity, denote the Bayes risk by $R\left( \hat{\theta}(\cdot) \right) := R_{\Omega}\left( \hat{\theta}(\cdot) \right)$ and the Bayes-optimal risk by $R^\star :=  R(\hat{\theta}^\star)$.
Moreover, let $\mathcal F_m$ be the class of DeepSets networks with capacity parameter (embedding dimension) $m$. 
For a training sample of size $N$ consisting of simulated pairs $(\theta_i,\mathcal{X}_i)\overset{iid}{\sim}\pi(\theta) f(\cdot\mid\theta)$ define the empirical (Monte--Carlo) risk
\[
 R_N(\hat\theta) \;=\; \frac{1}{N}\sum_{i=1}^N L(\theta_i,\hat\theta(\mathcal{X}_i)).
\]

Let $\hat \theta_{m,N}^\star, \hat\theta_m^\star \in\mathcal F_m$ denote then empirical and population risk minimizer over $\mathcal F_m$, respectively,
\[
\hat \theta_{m,N}^\star \in \arg\min_{\hat{\theta_m} \in \mathcal F_m}   R_N(\hat\theta_m) \quad,\quad \hat\theta_m^\star \in \arg\min_{\hat{\theta}_m\in\mathcal F_m} R(\hat\theta_m).
\]

We now establish the consistency result under the following assumptions.

\begin{enumerate}[label=(A\arabic*)]
  \item \textbf{Compactness, continuity and Lipschitz loss.} 
  The parameter space $\Theta$ is compact. 
  The loss $L(\theta,a)$ is continuous on $\Theta\times\Theta$ and is uniformly Lipschitz in its second argument, i.e., there exists $0<C_L<\infty$ such that
  \[
  \forall \theta\in\Theta,\ \forall a,a'\in\Theta,\qquad
  |L(\theta,a)-L(\theta,a')|\le C_L\|a-a'\|.
  \]
  
  \item \textbf{Permutation invariance of Bayes rule.} For i.i.d.~replicates $\mathcal{X}=(X_1,\dots,X_n)$ the Bayes estimator $\hat\theta^\star(X)$ is a permutation--invariant function of the sample.

  \item \textbf{DeepSets universality.} The union $\bigcup_{m\ge1}\mathcal F_m$ is dense in the space of continuous permutation-invariant functions on the sample space \cite{zaheer2017deepSets}.
  This means that, for every continuous permutation-invariant $g$ on $\mathbbm{R}^n$ and every $\varepsilon>0$ there exists $m$ and $\hat\theta\in\mathcal F_m$ with $\sup_x\|g(x)-\hat\theta(x)\|<\varepsilon$.

  \item \textbf{Uniform law of large numbers.} For each fixed $m$, the class $\mathcal F_m$ satisfies,
  \[
  \sup_{\hat\theta\in\mathcal F_m} \big| \hat R_N(\hat\theta) - R(\hat\theta) \big| \xrightarrow{\,P\,} 0
  \quad\text{as } N\to\infty.
  \]
\end{enumerate}

\begin{theorem}  \label{th:risk.consistency}
Assume (A1)--(A4). 
Fix a sequence $m\to\infty$ and, for each fixed $m$, let $N\to\infty$.
For any sequence of empirical risk minimizers $\hat\theta_{m,N}^\star\in\mathcal F_m$, we have 
\[
  R(\hat \theta_{m,N}^\star) \xrightarrow{\,P\,} R^\star
\]
for any joint sequence $\min(m,N)\to\infty$ such that $N$ increases sufficiently rapidly relative to $m$ so that Assumption (A4) holds for each class $\mathcal F_m$. 
\end{theorem}

The proof of Theorem~\ref{th:risk.consistency} can be found in the Appendix (Section~\ref{sec:proofTheorem2} ).
This theorem shows that, under a conceptually simple set of assumptions, the risk of the neural Bayes estimator converges to the Bayes risk in the DeepSets framework, as both the network capacity $m$ and the simulation budget $N$ grow.

Although this result is novel in the context of DeepSets, related progress has been made on the theoretical guarantees of neural Bayes estimators.
In particular, the recent paper by \cite{rodder2025} develops a similar risk decomposition framework to establish convergence results for a broad class of neural networks architectures.
Their analysis is technically more involved and arguably more general, as it incorporates an explicit optimization error term.
However, our framework has the advantage of using a general loss function in our analysis.
Readers interested in sharper bounds, explicit treatment of optimization error, or a discussion of how fast $N$ must grow relative to $m$ are referred to \citep{rodder2025}, who explore these issues in depth.

If we further strengthen our assumptions to require strong convexity of the loss function, then the Bayes minimizer is unique, and we obtain a stronger conclusion. 
In particular, suppose that:
\begin{enumerate}[label=(A\arabic*)]\setcounter{enumi}{4}
  \item The loss function $L(\theta, a)$ is strictly convex in $a \in \Theta$.
  \item The Bayes risk $R_{\Omega}\left( \hat{\theta}(\cdot) \right)$ is finite with respect to the prior $\Omega$ chosen.
\end{enumerate}

Under these assumptions, the Bayes estimator $\hat{\theta}^\star \in \arg\min_{\hat{\theta}}  R_{\Omega}( \hat{\theta} )$ is uniquely defined (see \cite[Chapter 4, Cor.~1.4]{lehmann1998})).
Then, compactness of $\Theta$ and uniqueness of the minimizer imply that $\hat\theta_{m,N}^\star \to \theta^\star$ in probability.
The next corollary makes this statement precise.

\begin{corollary}
Assume (A1)--(A6) hold, for any sequence of empirical Bayes estimators $\hat\theta_{m,N}^\star \in \mathcal{F}_m$, we have
\[
\hat\theta_{m,N}^\star \xrightarrow{\,P\,} \theta^\star
\quad \text{as } \min\{m, N\} \to \infty,
\]
provided $N \to \infty$ sufficiently fast relative to $m$.
\end{corollary}

\section{Simulation Studies} \label{sec:simpleModels}

With the overall estimation framework established, we test the proposed framework on a set of standard L\'evy process models, as well as more recent multiply-subordinated options. 
For each model, we compare against extant estimation procedures. 

\subsection{Experimental Setup}

Before proceeding, there are various design choices in the NBE setup that must be chosen, such as the number of training samples, neural network architecture, loss function, etc. 
Based on Theorem ~\ref{theo:Levy}, we can, in principle, generate as many datasets as needed, so $K$ and $J$ could theoretically be chosen arbitrarily large. 
In practice, however, the amount of training data required depends on several factors, including the complexity of the model, the number of parameters being estimated, the chosen loss function, the neural network architecture, and even the specifics of the optimization algorithm. 
As a rough guideline, \citet{sainsbury2023nbe} suggest that values of $K$ on the order of $10^4$–$10^5$ are typically necessary for the neural Bayes estimator to generalize well.
Our simulation experiments are set up to guide the selection of these components. 
Full details of these simulation studies are provided in the Appendix (Section~\ref{sec:appSimulations}).

The results we present in this section are obtained using fully connected dense neural networks (DNNs) for both the summary network $\Lambda$ and the inference network $\phi$. 
Each network consists of 3 hidden layers with 32 neurons per layer. 
Although we explored a variety of alternative architectures to support this choice, it is worth noting the performance of NBE appears generally robust to moderate changes in network design, as observed in our simulation studies.

Based on the results from our tuning experiments, we selected the mean as the aggregation function, Leaky ReLU as the activation function, and the mean squared logarithmic error (MSLE) as the loss function. 
These choices provided the best overall performance across the different parameters considered.
Finally, we assume that the model parameters are independent a priori and place uninformative priors on each, using uniform distributions over fixed intervals. 
These choices strike a balance between accuracy and computational efficiency, enabling us to focus on evaluating the estimator’s performance across different L\'evy process models. 

In the following analyses, as well as in the simulation studies, all methods were implemented in \texttt{Julia} to ensure comparability of computational timings.
We implement the NBE approach using the \texttt{NeuralEstimators} package \citep{nbePackage}. 
The LSQ estimator was coded using the \texttt{Optim.jl} package \citep{mogensen2018}, specifically calling the \texttt{optimize} function. 
The MELE estimator was implemented using \texttt{JuMP.jl} \citep{dunning2017} with the \texttt{Ipopt} solver \citep{wachter2006}, where the empirical likelihood was directly maximized. 
Furthermore, all experiments were conducted on an Apple M3 MacBook Pro with a 10-core CPU and 18 GB of unified memory.
The remainder of this section describes the parameter settings, simulation structure, and comparative benchmarks used in our evaluation.

\subsection{Standard L\'evy Models}\label{sec:simpleModels}

To explore performance of the proposed approach, we compare our method against two established estimation approaches: the LSQ method used in \citep{dvgBerry2023} and the MELE method (see Section \ref{sec:classicalMethods}). 
For both, we used the same grid of characteristic function frequencies. 

We evaluate the performance of the proposed estimator on three benchmark models: the compound Poisson process, the variance gamma process, and the Merton jump-diffusion model. See Section \ref{sec:appSimulations} for a detailed description of the prior distributions for each model. 
For each model, we train the neural Bayes estimator and assess its performance on an independent test set of $1000$ datasets in which the goal is to estimate the model parameters based on simulated test data. 

Tables \ref{tab:resPoisson}, \ref{tab:resMerton}, and \ref{tab:resVarGamma} contain the average root mean squared error (RMSE), bias, and standard deviation for each parameter across the $1000$ test datasets, along with the total estimation time required by each method. 
Across all models and parameters, NBE consistently outperforms competing methods in terms of estimation accuracy, sometimes substantially. 
It achieves the lowest RMSE for every parameter in every model considered. 
Additionally, NBE demonstrates a significant advantage in computational efficiency. 
While traditional methods such as LSQ and MELE require several hundred seconds -- and in some cases over a hundred thousand seconds -- to process the full assessment set, NBE achieves comparable or better accuracy in just 3 to 6 seconds per model. 
This dramatic speed-up highlights the practical benefits of amortized inference: once trained, the neural Bayes estimator delivers high-quality parameter estimates with minimal computational cost.

It is important to note that the estimation times reported in Tables~\ref{tab:resPoisson}--\ref{tab:resVarGamma} refer exclusively to inference time. 
That is, the time required to produce parameter estimates once the neural Bayes estimator has already been trained. 
We emphasize that NBE involves a non-trivial upfront computational cost for training the network, whereas methods such as LSQ or MELE directly optimize on each dataset without training.
However, once the network has been trained, inference with NBE is essentially instantaneous, while LSQ and MELE require the solving of a new optimization problem for every data set.
This is a key advantage of our approach: once training is complete, inference can be performed extremely quickly. 
This characteristic is known as \textit{amortized learning}—a framework where computational cost is concentrated upfront during a training phase, after which model execution is fast.

While the initial training phase can be computationally intensive, it only needs to be performed once per model specification and prior distribution. 
In our own experiments, training time varied depending on the complexity of the model and the choices of hyperparameters. 
Across all simulation settings, we observed training durations ranging from approximately 2 hours for simpler models to as much as 55 hours for more complex configurations, such as the level-2 DVG process with large $K$ and $n_t$. 
On average, training required around 25 hours. 
However, these training costs are offset by the significant gains in speed and scalability at inference time, which is particularly beneficial for high-frequency financial data.

\begin{table}[t]
    \centering
    \resizebox{\textwidth}{!}{%
    \begin{tabular}{|c|c|c|c|c|c|}
      \hline
         Compound Poisson process  & Training [s] & Est Time[s] &  $\lambda$ & 
      $\mu$ & $\sigma^2$ \\
      \hline 
      LSQ  & - & 833 & 0.058 (-0.004) [0.038] & 0.23 (-0.025) [0.21] & 0.13 (0.116) [0.065] \\
      MELE & - & 777.7 [h]* & 0.70 (-0.61) [0.38] & 0.36 (-0.038) [0.19] & 0.11 (-0.09) [0.067] \\
      NBE & 1276 &  \textbf{2.8} & \textbf{0.051} (\textbf{0.0005}) [\textbf{0.032}] & 
      \textbf{0.028}(\textbf{-0.0055})[\textbf{0.018}] & \textbf{0.022}(\textbf{7.6e-5}) [\textbf{0.014}] \\
      \hline
    \end{tabular}%
    }
    \caption{RMSE, (Bias) and [SD] (averaged over 1000 simulated datasets from the prior) for the jump intensity ($\lambda$), mean ($\mu$) and variance ($\sigma^2$) of each jump of a compound Poisson process. 
    The table also includes the time required to estimate the parameters over all simulations. 
    * For MELE, only 10 datasets were computed due to its high computational cost. The reported time is extrapolated from these 10 datasets.
    The results were obtained using $n_t=500$, $K=5000$ and $J=10$.}
    \label{tab:resPoisson}
\end{table}

\begin{table}[t]
    \centering
    \resizebox{\textwidth}{!}{%
    \begin{tabular}{|c|c|c|c|c|c|c|c|}
      \hline
        Merton model   & Training[s] & Est Time[s] &  $\mu$ & 
       $\sigma^2$ &  $\lambda$ &  $\mu_j$&  $\sigma^2_j$ \\
      \hline 
      LSQ  & - & 695 & 0.17 (-0.001) [0.14] & 0.19 (0.14) [0.091] & 0.46 (0.13) [0.34] & 0.34 (\textbf{0.007}) [0.25] & 0.37 (-0.056) [0.25] \\
      MELE & - & 1155 [h] * & 0.37 (-0.21) [0.27]  & 0.20 (-0.12) [0.11] & 1.01 (0.43) [0.39] &  0.31 (-0.14) [\textbf{0.083}]& 0.84 (-0.69) [0.53] \\
    NBE &  2085 & \textbf{4} & \textbf{0.11} (\textbf{4.5e-5}) [\textbf{0.072}]& 
      \textbf{0.11} (\textbf{-0.004}) [\textbf{0.068}]& \textbf{0.25} (\textbf{0.022}) [\textbf{0.16}] &\textbf{0.19} (-0.09) [0.13] & \textbf{0.24}(\textbf{-0.015}) [\textbf{0.15}]  \\
      \hline
    \end{tabular}
    }
    \caption{RMSE, (Bias) and [SD]  (averaged over 1000 simulated datasets from the prior) for the mean ($\mu$) and variance ($\sigma^2$) of the diffusion component, along  with the jump intensity ($\lambda$), mean ($\mu_j$) and variance ($\sigma_j^2$) of the jump component of a Merton model. 
    The table also includes the time required to estimate the parameters for the 1,000 simulated datasets.
    * For MELE, only 10 datasets were computed due to its high computational cost. The reported time is extrapolated from these 10 datasets.
    The results were obtained using $n_t=1000$, $K=5000$ and $J=10$.}
    \label{tab:resMerton}
\end{table}

\begin{table}[t]
    \centering
    \resizebox{\textwidth}{!}{%
    \begin{tabular}{|c|c|c|c|c|c|}
      \hline
        Variance gamma process   & Training [s] & Est Time[s] &  $\gamma$ & 
       $\sigma^2$ &  $\alpha$ \\
      \hline 
      LSQ  & - & 398 & 0.16 (\textbf{-0.005}) [0.15]& 0.32 (-0.083) [0.21]& 1.70 (-0.59) [0.85]\\
      MELE & - &  1024[h] * & 0.73 (-0.025) [0.40] & 1.18 (-0.99) [0.69]& 1.93 (-0.98) [0.37]\\
    NBE &  672 &  \textbf{3.5} & \textbf{0.048} (-0.01) [\textbf{0.030}]& 
      \textbf{0.090} (\textbf{-0.0085}) [\textbf{0.062}]& \textbf{0.24} (\textbf{0.038}) [\textbf{0.17}]\\
      \hline
    \end{tabular}
    }
    \caption{RMSE, (Bias) and [SD]  (averaged over 1000 simulated datasets from the prior) for the drift ($\gamma$), variance ($\sigma^2$) and variance ($\alpha$) of the subordinator for a variance gamma process.
    The table also includes the time required to estimate the parameters for the 1,000 simulated datasets.
    * For MELE, only 10 datasets were computed due to its high computational cost. The reported time is extrapolated from these 10 datasets.
    The results were obtained using $n_t=1000$, $K=5000$ and $J=10$.}
    \label{tab:resVarGamma}
\end{table}

\subsection{Deep Variance Gamma} \label{sec:simDVG}

We test our framework on a more difficult estimation task involving multiply-subordinated processes; in particular we consider a two-level DVG model. 
As outlined in Section \ref{sec:dvg}, the two-level DVG parameters are the variance parameter $\sigma^2$ and the subordination parameters $\alpha_1, \alpha_2$, corresponding to each level of gamma subordination. 

Combining the recommendations of \citet{sainsbury2023nbe} and our simulation studies (see Appendix~\ref{sec:appSimulations}), we fix $K = 10,000$ parameter samples and $J = 10$ datasets per parameter, yielding a training set of $100,000$ simulations (see Equation (\ref{eq:MCApprox})). 
The input dimension $n_t$, the number of observations passed to the network, can be application dependent. 
As our applied interest is in one-minute cryptocurrency returns over individual days, we use an input size of $n_t = 1440$, i.e., a day worth of data.
Although this is easily changed for other applications, it does involve retraining the network.

In terms of the architecture used, we used the same one as in the previous section and the same aggregation, activation, and loss function. 
Finally, we trained the neural Bayes estimator and evaluated its performance on 1000 datasets, each generated from parameter values sampled from the prior distribution.

We selected prior distributions to be reasonably uninformative, reflecting limited prior knowledge while remaining consistent with ranges typically used in empirical studies of financial L\'evy models and allowing for sufficient flexibility in the estimation, especially when applied to multiple datasets. 
For instance, variance parameters in L\'evy models are often on the order of 1 or smaller, and subordination parameters typically fall within moderate positive ranges \citep{dvgBerry2023}. 
Accordingly, we chose uniform priors: $\sigma^2 \sim U(10^{-6}, 3)$ and $\alpha_1, \alpha_2 \sim U(10^{-6},25)$, allowing the model to explore a broad but plausible range of parameter values.  
These bounds are informed by \citet{dvgBerry2023}, who showed that cryptocurrency parameter estimates consistently fell within these intervals, so the truncation reflects a combination of domain expertise and empirical evidence. 
Despite this, we acknowledge that even flat priors can exert influence. 
For general discussion of this issue see \citet{gelman2017}, and for a robustness experiment in our NBE setting, we refer the reader to Section ~\ref{sec:appSimulations}.

\citet{dvgBerry2023} also showed that potential model identifiability concerns may arise in the DVG, with certain parameter combinations exhibiting very similar implied characteristic functions, except for at either very small, or very large, frequencies. 
Thus, to assess the quality of our estimates, we introduce an additional metric that captures functional differences between the estimated and true models. 
Given the true parameter vector $\theta=\left( \sigma^2, \alpha_1, \alpha_2 \right)\T$, we compute the associated theoretical characteristic function $\Phi(\omega,\theta)$ at frequency $\omega$. 
Likewise, for an estimate $\hat{\theta}=\left( \hat{\sigma}^2, \hat{\alpha}_1, \hat{\alpha}_2 \right)\T$, we compute the corresponding $\Phi(\omega,\hat{\theta})$. 
We define the $\mathbb{L}_2^{f}$ distance between these vectors as 
\begin{equation}
    \| \theta-\hat{\theta} \|_{\mathbb{L}_2^{f}} = \displaystyle \sqrt{\displaystyle \int_{\omega} \left(\Phi(\omega,\theta)-\Phi(\omega,\hat{\theta})\right)^2 d\omega}
\end{equation}
which serves as a functional discrepancy measure between the true and estimated characteristic functions. 
This metric provides a complementary perspective to RMSE by quantifying the difference in the implied distributions, and it helps to evaluate whether small parameter errors lead to meaningful deviations in the underlying process behavior.
  
\begin{figure}[t]
 \centering
    \includegraphics[width=\textwidth]{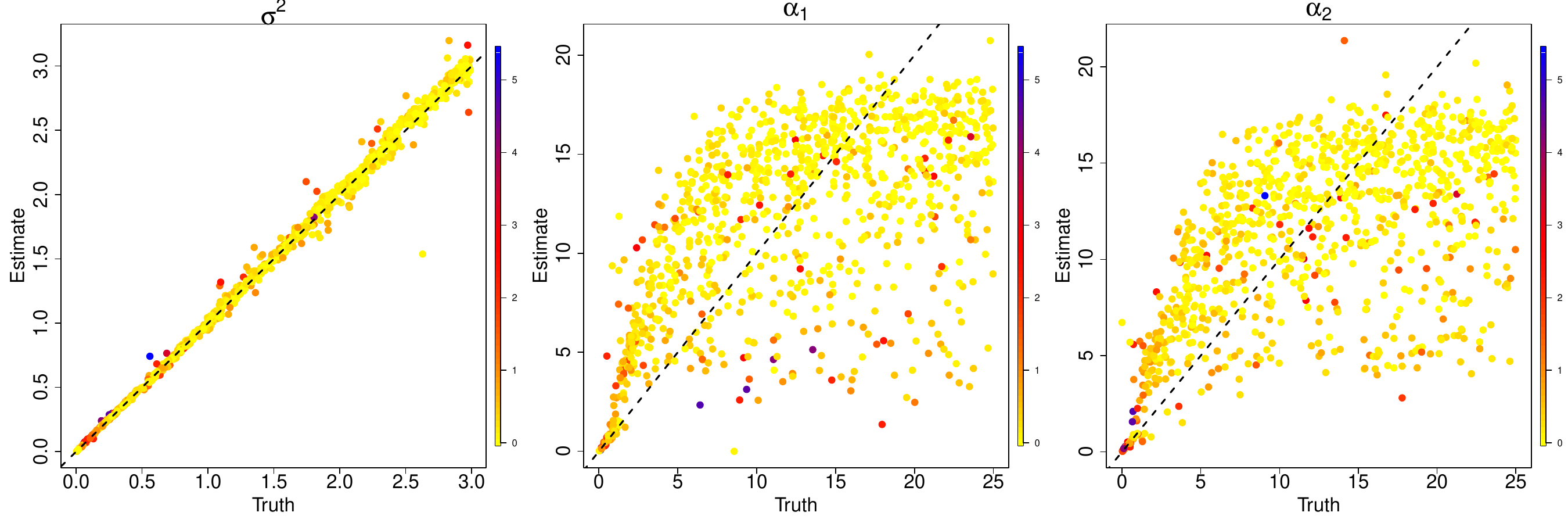}
  \caption{Scatter plots of true versus estimated values for $\sigma^2, \alpha_1$, and $ \alpha_2$, colored by their $\mathbb{L}_2^{f}$ difference.
  The dashed black line represents the identity.}
    \label{fig:l2phi}
\end{figure}

Using the $\mathbb{L}_2^{f}$ metric introduced above, Figure \ref{fig:l2phi} presents a scatterplot of the true versus estimated values for $\sigma^2, \alpha_1$, and $\alpha_2$, respectively. 
Each point is colored according to its $\mathbb{L}_2^{f}$ discrepancy, providing a visual cue for the functional accuracy of the estimator. 
The results show that the model performs well in estimating the variance $\sigma^2$, with estimates closely aligned with the identity line. 
However, estimation of the subordinator parameters $\alpha_1$ and $ \alpha_2$ appears more challenging, particularly in cases where their values are small. 

It should be noted that Bayes estimators are, in general, biased and therefore are not necessarily expected to lie on the identity line in Figure~\ref{fig:l2phi}. 
In principle, we could assess the role of this bias by comparing our neural Bayes estimates with the corresponding true Bayes estimates. 
However, computing the exact Bayes benchmarks is not feasible for the class of models considered here, which is precisely why we adopt a likelihood-free approach. 

Despite this, the color scale reveals that even when the parameter estimates deviate from their true values, the resulting characteristic function remains close to the true one. 
This suggests that the model still captures the key structural properties of the data, highlighting its robustness in functional inference even when parameter-level accuracy is imperfect.

\begin{figure}[t]
  \centering
  \includegraphics[width=\textwidth]{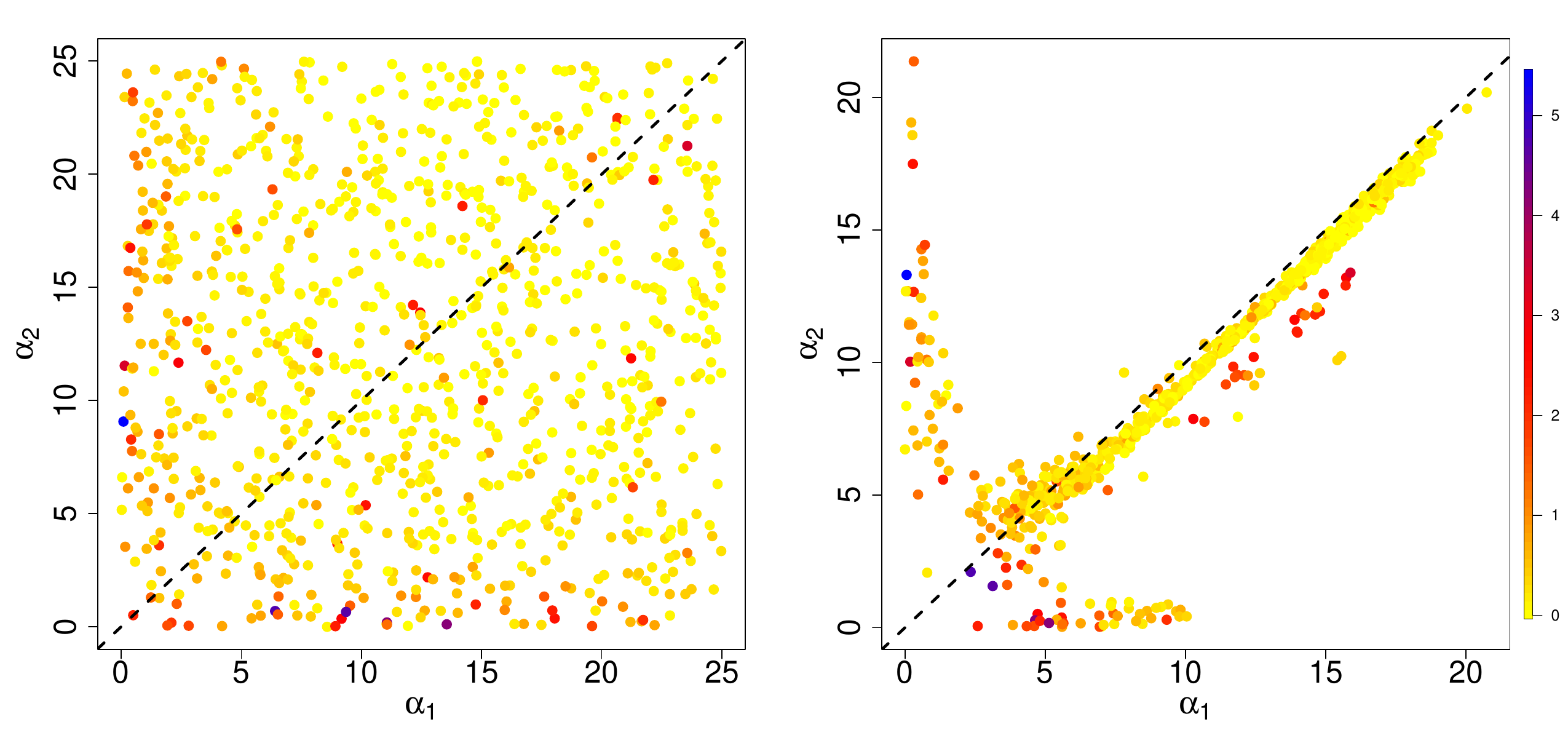}
  \caption{Scatter plots of $\alpha_1$ versus $\alpha_2$, colored by their $\mathbb{L}_2^{f}$ difference. 
  The left panel shows true parameters, while the right panel shows the corresponding estimates.}
  \label{fig:subParam}
\end{figure}

Interestingly, the results also suggest a strong correlation between the two subordinator parameters, which prompted further investigation. 
Figure \ref{fig:subParam} displays scatter plots of $\alpha_1$ versus $\alpha_2$, both for the true parameter values and the estimates. 
Several insights emerge from this visualization. 
First, estimation appears to be more difficult when either $\alpha_1$ or $\alpha_2$ is close to zero, as indicated by higher $\mathbb{L}_2^{f}$ values near the bottom an left edges of the plot.
More notably, the estimated values show a tendency for the model to predict $\alpha_1$ and $\alpha_2$ as being similar. 
This alignment suggests that the neural Bayes estimator finds its best functional fit when $\alpha_1 \approx \alpha_2$, potentially revealing a structural symmetry in the underlying process.

\section{Application to Cryptocurrency Returns}\label{sec:Results}

Cryptocurrencies have become increasingly popular within the last decade. 
Led by Bitcoin (BTC), the market comprises hundreds of blockchain-based assets, with Ethereum (ETH) and Ripple (XRP) among the most prominent.

Recent work has explored the use of L\'evy processes to model cryptocurrency returns. For example, \citet{shirvani2024} used a doubly subordinated normal inverse Gaussian model, and \citet{dvgBerry2023} applied deep variance gamma models to high-frequency data. 
Other studies have fit L\'evy stable distributions \citep{kakinaka2020} or  L\'evy-GJR-GARCH models \citep{wu2025}. 
While these approaches offer valuable insights, they are often limited to short time spans due to the high cost of parameter estimation.

In contrast, the neural Bayes estimation framework we employ enables fast, scalable parameter estimation. 
This makes it possible to analyze large-scale, high-frequency datasets spanning long time horizons—such as an entire year of minute-by-minute cryptocurrency returns—with minimal computational burden.

For our analysis, we obtained high-frequency 1-minute price data from \citep{kaggleCrypto}. 
For each currency, we considered data spanning the calendar year of 2022. 
While some currencies contain some gaps, we address this by resampling missing values from the observed log-returns, under the assumption that log-returns are stationary and independent. 
\citet{dvgBerry2023} showed that DVG models are suitable for high-frequency cryptocurrency data; thus we model the log-returns using a level-2 DVG L\'evy process, and examine how the estimated parameters evolve over time for each currency.

\begin{figure}[t]
 \centering
    \includegraphics[width=\textwidth]{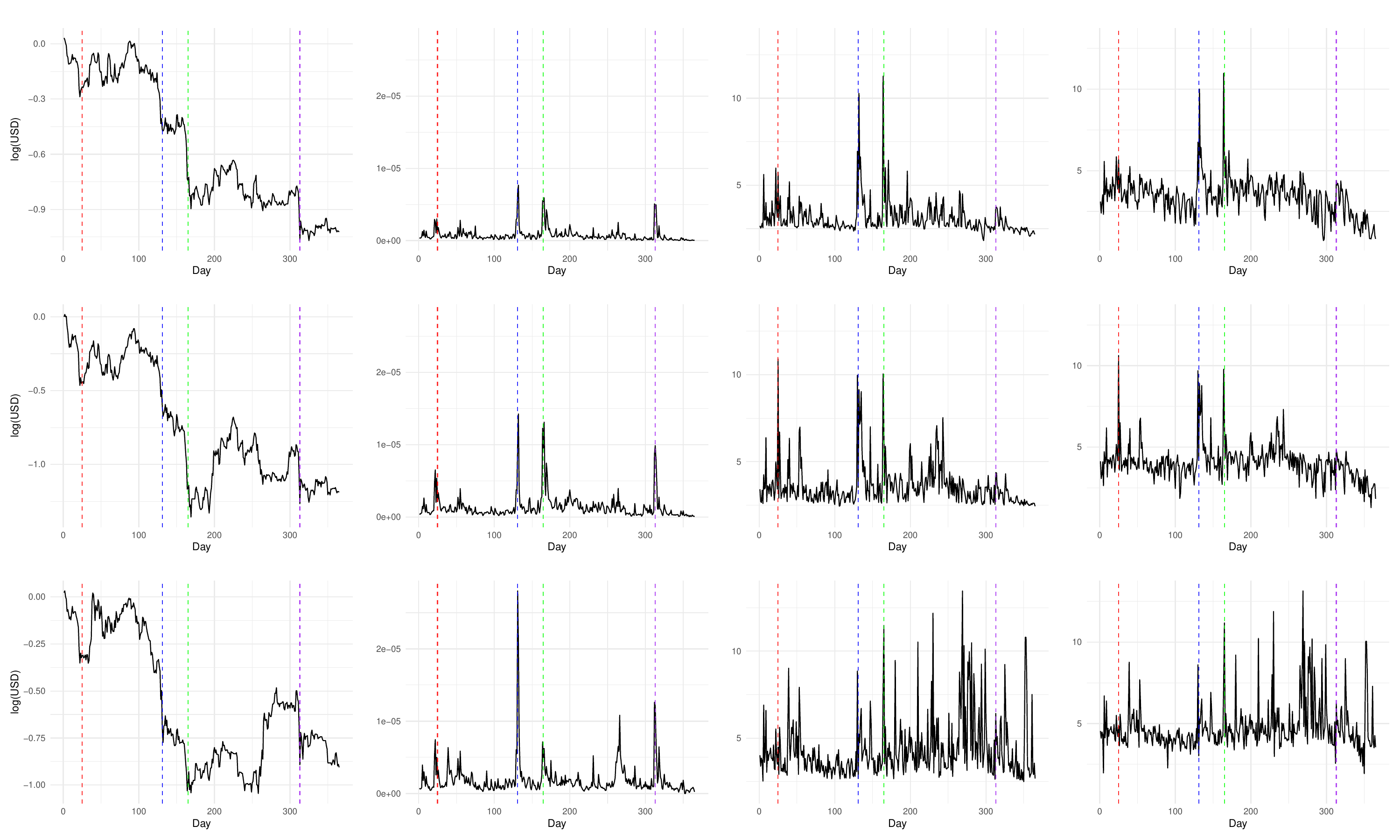}
  \caption{Daily parameter estimation using NBE for 2022. Each row represents bitcoin (BTC), ethereum (EHT) and ripple (XRP) respectively. The left most panel represents the log returns data. The following panels are the estimates of $\sigma^2$, $\alpha_1$, and $\alpha_2$, of a level $2$ DVG process, respectively.  The dashed lines corresponds to Jan 25, May 11, Jun 14, and Nov 11.}
    \label{fig:estParam}
\end{figure}

\begin{figure}[t]
 \centering
    \includegraphics[width=\textwidth]{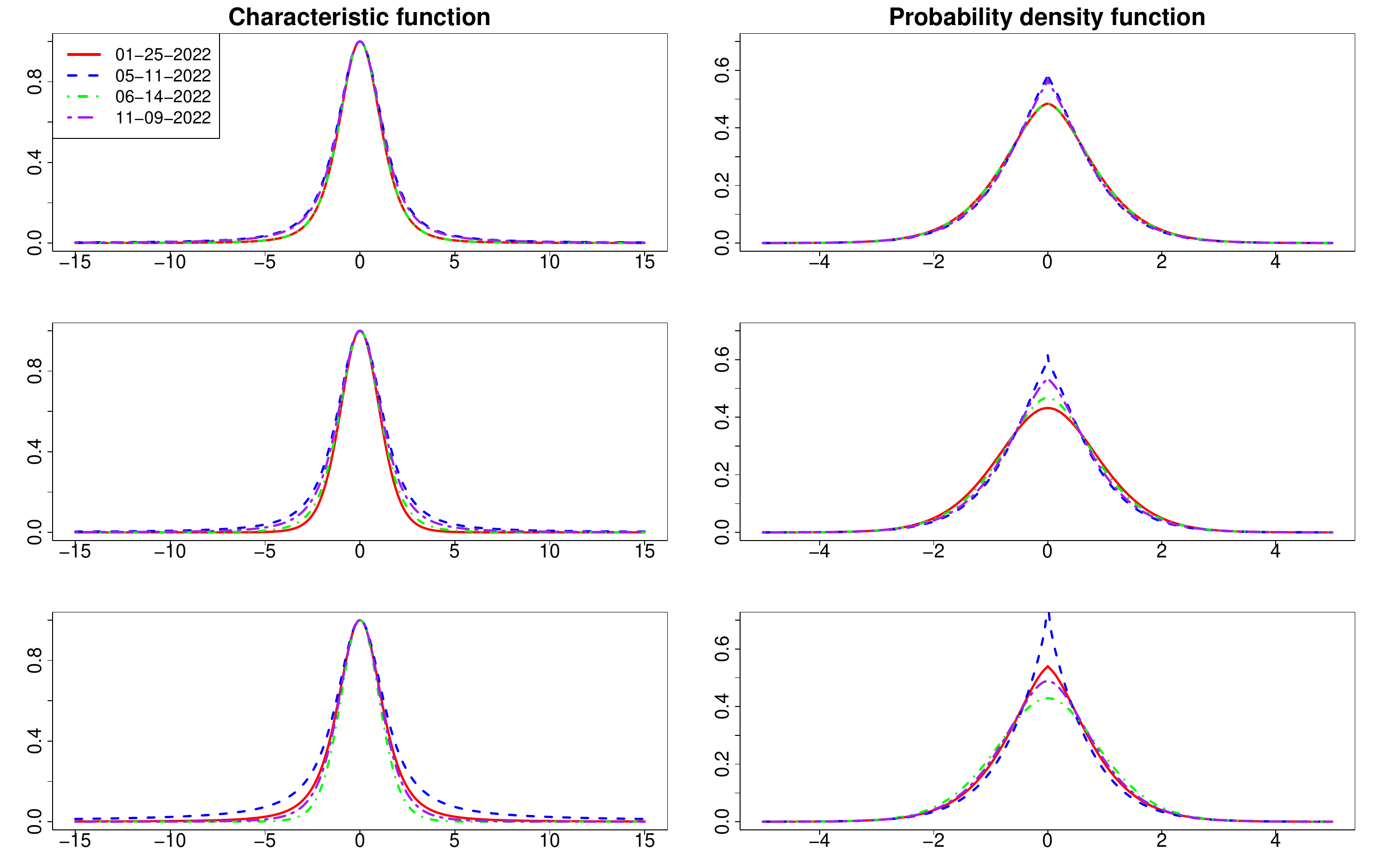}
  \caption{Estimated characteristic functions (left column) and probability density functions (right column) for selected days in 2022, using estimated parameters.
  Each row corresponds to a different cryptocurrency: BTC (top), ETH (middle), and XRP (bottom).}
    \label{fig:chfpdf}
\end{figure}


\begin{figure}[t]
 \centering
    \includegraphics[width=\textwidth]{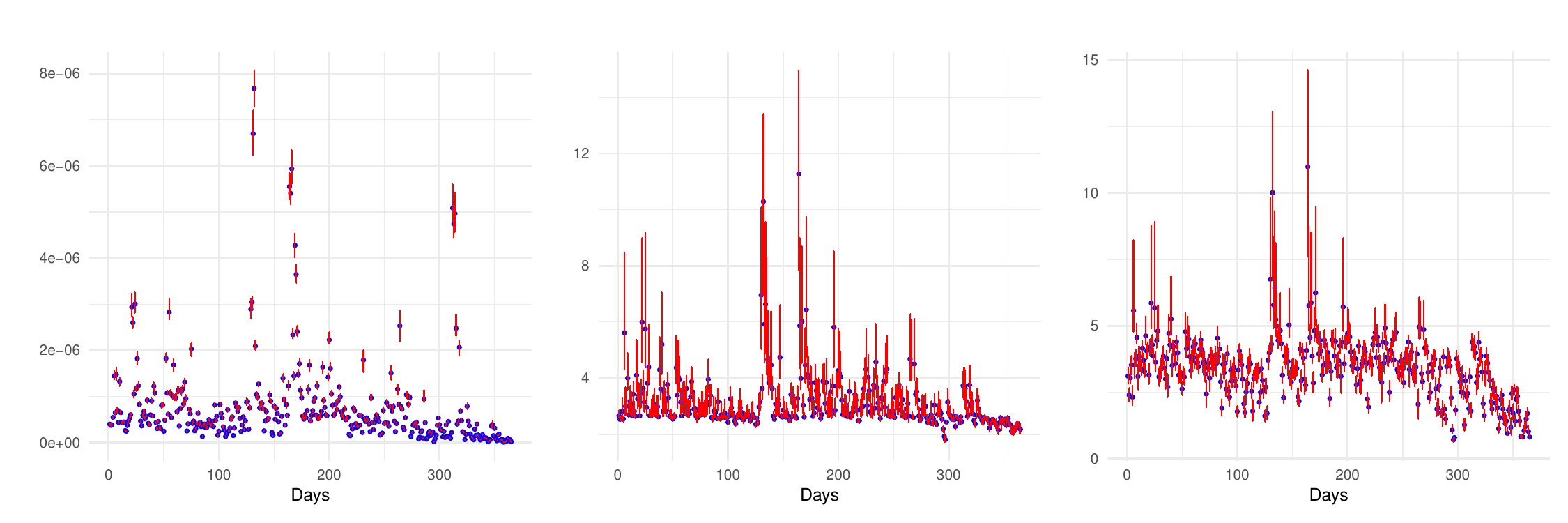}
    \caption{Estimated daily parameters and their $90\%$ confidence intervals for of a level-2 DVG model for BTC log returns in 2022.}
    \label{fig:UQparamNBE}
\end{figure}

\begin{figure}[t]
 \centering
    \includegraphics[width=\textwidth]{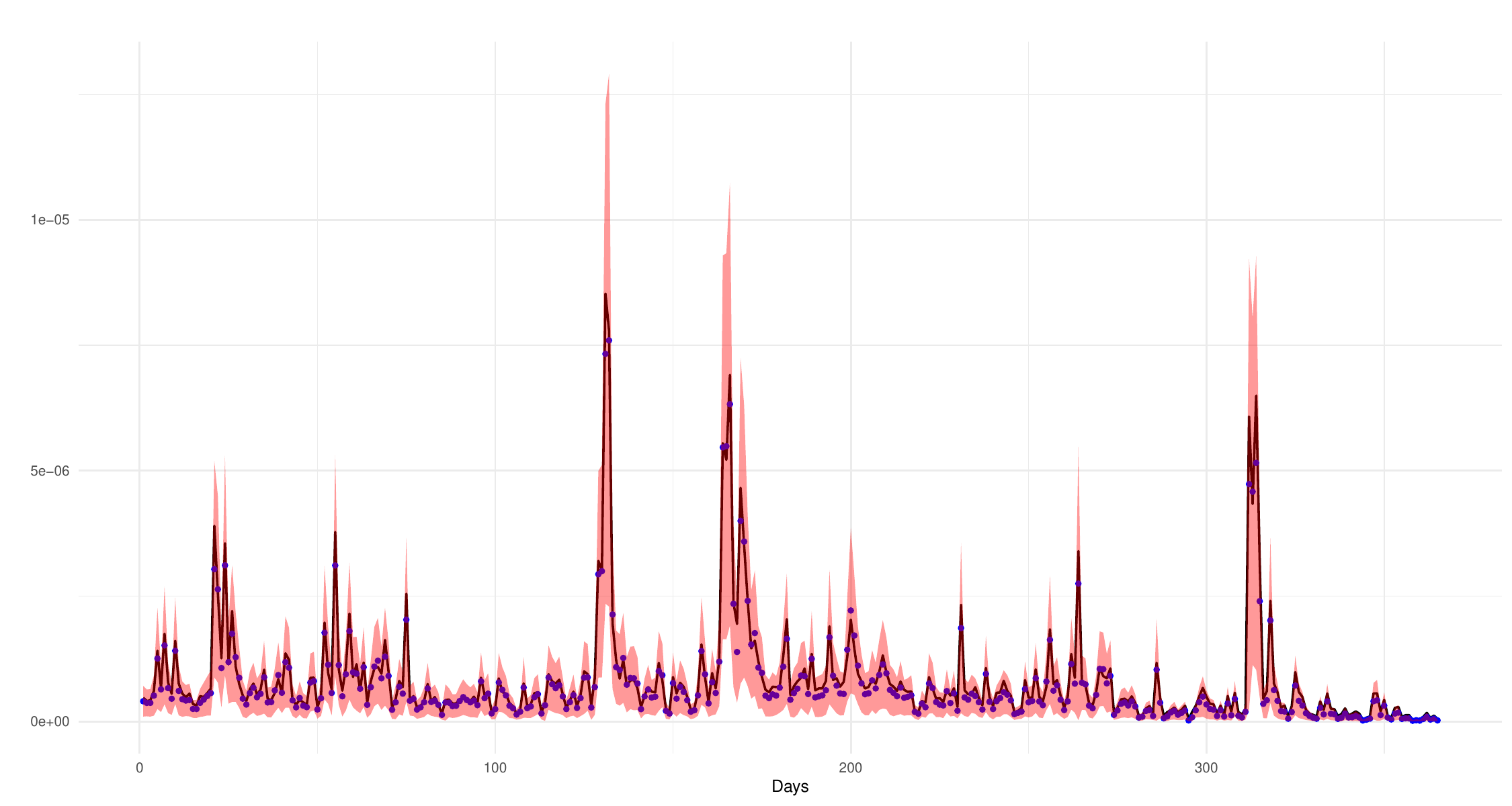}
    \caption{Estimated daily parameters of a level-2 DVG model for BTC log returns in 2022, with $90\%$ credible intervals obtained from the posterior distribution. 
The black line shows the posterior mean, while the blue dots indicate the posterior median.}
    \label{fig:UQlinlin}
\end{figure}


Figure \ref{fig:estParam} presents daily parameter estimates obtained using neural Bayes estimation for BTC, ETH, and XRP. 
The leftmost panel in each row shows the log-return time series, followed by the corresponding estimates for $\sigma^2$, $\alpha_1$, and $\alpha_2$. 
Notably, the model captures sharp increases in variance on days with significant market jumps, highlighting its responsiveness to volatility. 
Four vertical dashed lines mark key dates—January 25, May 11, June 14, and November 11 -- each associated with a major market event or crash.

It is worth noting that, in our studies, we found that the empirical variance of daily log returns was often quite small. Given that our prior for $\sigma^2$ was uniform on the interval $(10^{-6}, 3)$ and that only $10,000$ parameter samples were drawn, the lower end of the variance distribution was underrepresented during training. 
This made accurate estimation more difficult, particularly when comparing characteristic and density functions.
To address this, we applied a simple but effective rescaling strategy: for each day, we normalized the observed data by dividing it by its empirical standard deviation, effectively centering the variance around one. 
After obtaining parameter estimates from the neural network, we then re-scaled the estimated variance by multiplying it with the original empirical standard deviation. 
This preprocessing step improved stability and provided more sensible results in low-variance regimes.

While BTC, ETH, and XRP share general trends in their price dynamics, XRP exhibits notably higher volatility in its estimated parameters—consistent with the findings of \citet{celeste2020}, who report that Bitcoin exhibits smoother volatility dynamics over time compared to Ripple and Ethereum, suggesting a higher degree of market maturity for BTC relative to other cryptocurrencies.
We conjecture that this discrepancy likely stems from structural differences among the assets. 
Bitcoin, as the most established and decentralized cryptocurrency, benefits from deeper liquidity, greater institutional adoption, and a broader investor base. 
XRP, by contrast, is subject to centralization concerns, ongoing regulatory scrutiny, and a more speculative trading environment -- factors that, according to \citet{celeste2020}, ``suggest that price dynamics may be instead dominated by short-termism and crowd behavior.''

Figure \ref{fig:chfpdf} provides further insight by comparing characteristic functions and probability density functions for the selected high-volatility days. 
These plots illustrate the flexibility of the NBE approach in adapting to changing market conditions. 
The day-to-day variation in the characteristic and density functions underscores the nonstationary nature of financial data, and how the model dynamically adjusts to capture this complexity.

Among these dates, May 11, 2022, stands out. 
On this day, the cryptocurrency market experienced significant turmoil due to the collapse of the Terra blockchain ecosystem  \citep{liu2023TERRA}, specifically its algorithmic stablecoin UST and its companion token LUNA.  
This event sparked widespread panic across the crypto space. 
While Bitcoin and Ethereum registered large drops, their reactions were relatively smooth due to deeper liquidity and stronger investor confidence. 
In contrast, XRP, already under regulatory pressure from the SEC, displayed a more erratic response. 
This is reflected in our estimates by a more sharply peaked probability density function, suggesting a surge in small, frequent jump-like movements rather than a single large drop. 
Such behavior is consistent with panic-driven trading in low-liquidity environments, where prices fluctuate rapidly in response to uncertainty and fragmented information.

Figure \ref{fig:UQparamNBE} offers a visual representation of the $90\%$ confidence intervals that surround the parameter estimates. 
Notably, the non-overlapping confidence intervals throughout the year indicate that the underlying process is indeed nonstationary -- its parameters evolve meaningfully over time. 

While the bootstrap procedure provides estimator uncertainty, the NBE framework also enables uncertainty quantification in a fully Bayesian manner. 
Figure~\ref{fig:UQlinlin} shows the estimates for the posterior mean, median, and illustrates the $90\%$ posterior-based credible intervals for the process variance of Bitcoin in 2022. 
Comparing Figures~\ref{fig:UQparamNBE} and \ref{fig:UQlinlin}, we observe that the posterior credible intervals are generally wider than the bootstrap intervals, as they reflect full distributional uncertainty. 
This is particularly evident during periods of high volatility, where the posterior approach tends to be more conservative, while the bootstrap intervals remain relatively narrow in regions where the estimator is stable. 

In particular, we define two cryptocurrencies as having different distributions on a given day if at least one of their corresponding confidence intervals does not overlap.
Applying this criterion, we find that Bitcoin and XRP exhibit different distributions on 239 days in 2022, corresponding to roughly $65\%$ of the year.
Similarly, Bitcoin and Ethereum differ on approximately $40\%$ of the days, and Ethereum and XRP differ on about $38\%$ of the days.

\begin{table}[t]
    \centering
    \begin{tabular}{|c|ccc|}
      \hline
          & MELE & LSQ &  NBE \\
      \hline
      Time [s] & 45,552 & 656 & \textbf{5} (10.8 [hr]*)  \\
      \hline
    \end{tabular}
    \caption{Time to perform daily parameter estimation  for the year 2022 averaged across 3 different cryptocurrencies.
    * Training time for NBE. The reported 5 seconds refer to inference after training.}
    \label{tab:estimationTimes}
\end{table}

While these comparisons are exploratory in nature (and we do not formally correct for multiple testing; interested readers may consider adjustments such as the Bonferroni correction), the results nevertheless suggest persistent differences in the return behavior of these assets.
These findings are in line once again with \citet{celeste2020}, who reports that their method was able to ``provide interesting evidence that allows us to draw the conclusion that Bitcoin (BTC), Ethereum (ETH) and Ripple (XRP) may show quite different price behaviors.''

In addition to accuracy and robustness, computational efficiency is a key consideration for practical deployment. 
Table \ref{tab:estimationTimes} compares the estimation runtime of NBE against MELE and LSQ. 
NBE is approximately 9,000 times faster than MELE and 130 times faster than LSQ, making it an appealing option for real-time applications.

In summary, the NBE method demonstrates a strong ability to capture parameter dynamics in high-frequency cryptocurrency data. 
It delivers accurate estimates, provides robust uncertainty quantification, and does so with exceptional computational speed. 
These features position NBE as a powerful tool for modeling and analyzing complex financial systems.

\subsection{The Long-Term Evolution of Bitcoin}

\begin{figure}[t]
 \centering
    \includegraphics[width=\textwidth]{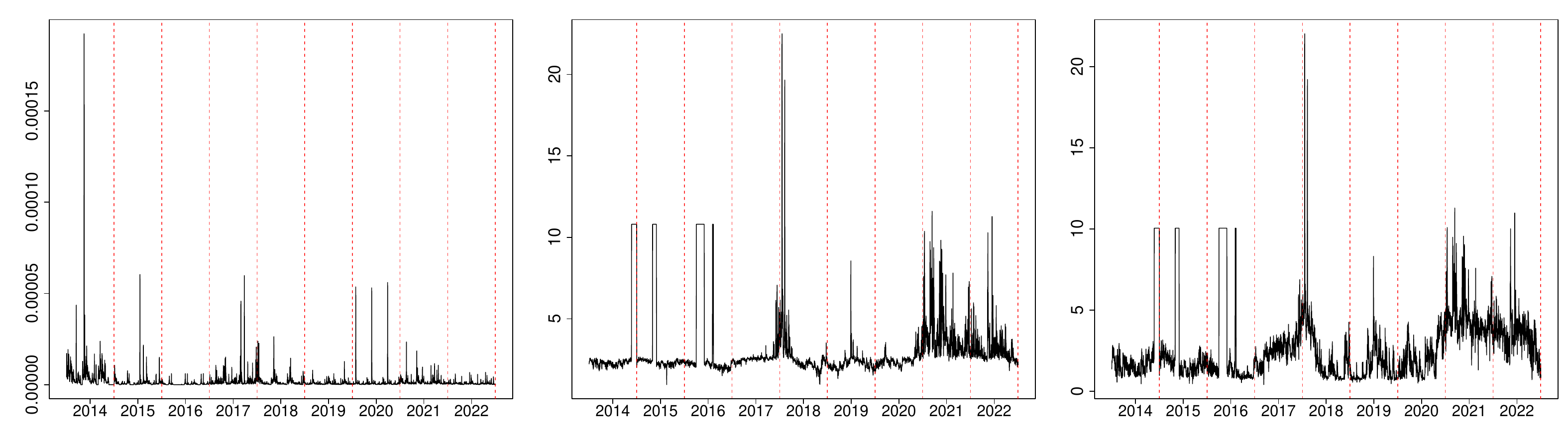}
    \caption{Estimated daily parameters $\sigma^2$, $\alpha_1$, and $\alpha_2$ for BTC log returns from 2014 to 2022 under the level-2 DVG model.}
    \label{fig:btc9yrParamEvo}
\end{figure}

\begin{figure}[t]
 \centering
    \includegraphics[width=\textwidth]{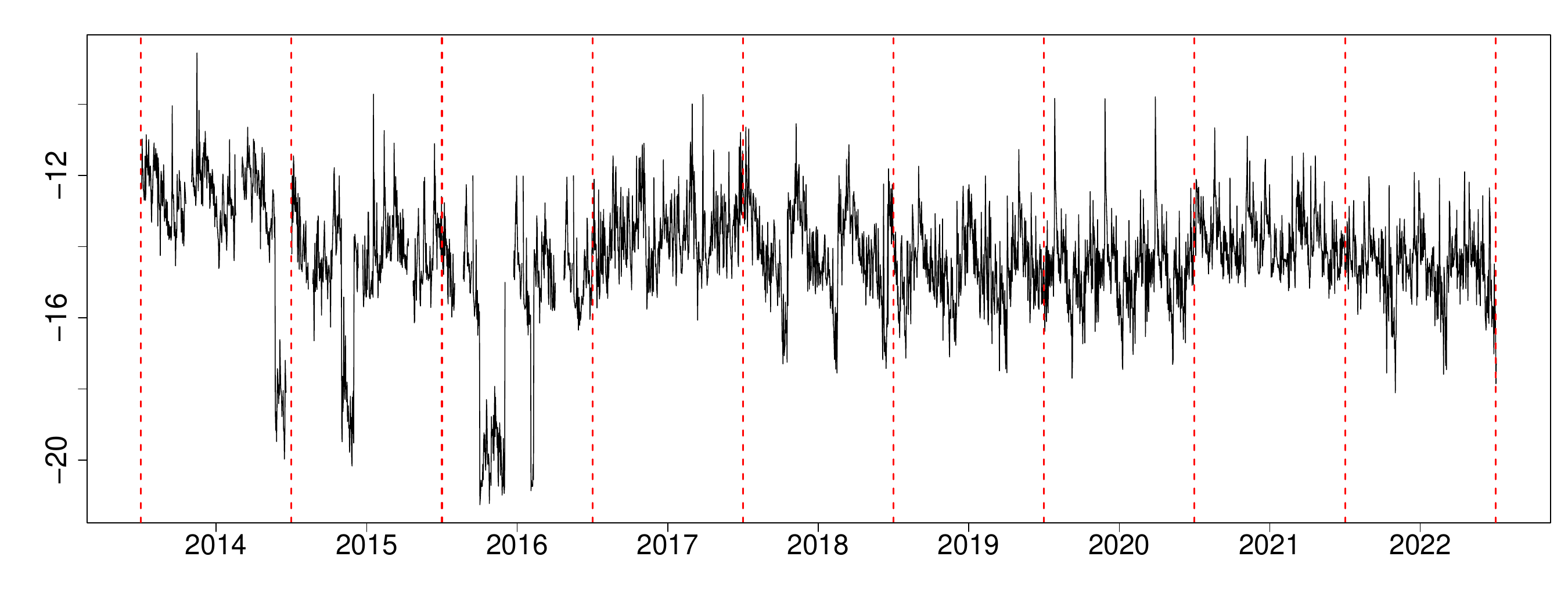}
    \caption{Evolution of the estimated log variance, $\log(\hat{\sigma}^2)$, for BTC from 2014 to 2022.}
    \label{fig:btc9yrlogVarEvo}
\end{figure}

We also analyzed the long-term evolution of DVG process parameters fitted to BTC  over a nine-year period spanning 2014 to 2022. 
The NBE approach shines in this application, with nearly a decade of data requiring only approximate 45 seconds to compute daily estimates for the entire period. 
Figure~\ref{fig:btc9yrParamEvo} displays these daily estimates for $\sigma^2$, $\alpha_1$, and $\alpha_2$ using the NBE framework. 
The parameters show clear signs of nonstationarity, with periods of relative stability punctuated by bursts of volatility. 
Since the variance estimates are difficult to interpret directly due to scale differences, we additionally visualize their evolution in log-scale in Figure~\ref{fig:btc9yrlogVarEvo}. 

Some interesting insights can be gleaned from this analysis; for example, the pronounced periodicity in BTC volatility.
We used a periodogram to look for apparent harmonic behavior and found a significant harmonic component with an approximate cycle length of two months. 
This aligns with broader findings in the literature: for example, \citet{celeste2020} observed long-term cyclical persistency and anti-persistency behavior in BTC prices. 
While their focus was on price dynamics, our findings suggest that similar cyclical behavior may also be present in the underlying volatility structure.

Another insight is the presence of the ``crypto winter'' around January 2018, where anomalously high gamma process parameters are estimated. 
It’s also worth addressing the unusual behavior observed in the early years, particularly 2014 to 2016. 
During this period, the source data contained substantial gaps — 46\%, 30\%, and 44\% for each respective year.
While these gaps are far from trivial, it’s notable that the estimated parameters still show broad consistency with the trends seen in later years, albeit with some additional noise. 

\section{Discussion}\label{sec:Conclusions}

Accurate parameter estimation remains a central challenge in modern financial modeling, particularly in settings where the likelihood function is either unavailable or computationally intractable. 
In this work, we demonstrated the effectiveness of neural Bayes estimation (NBE), a simulation-based, likelihood-free inference method that we tailored for estimating parameters in L\'evy processes.

NBE is a flexible, amortized framework for parameter estimation, in which a neural network trained via simulation is used to approximate Bayes estimators, functionals that minimize the Bayes risk. 
This allows NBE to bypass the need for closed-form likelihoods, making it especially well-suited for complex models such as L\'evy processes, where traditional methods are usually limited by computational cost.

Our work demonstrates that NBE not only provides computational efficiency and accuracy, but also comes with rigorous theoretical guarantees, including convergence in risk.
Moreover, our simulation studies showed that NBE consistently achieves lower estimation error across all parameters compared to established techniques such as LSQ and MELE. 
Moreover, it does so with dramatically reduced computation times, achieving speedups of several orders of magnitude, making it a compelling option for large-scale or time-sensitive applications.

The choice between NBE and traditional methods such as LSQ or MELE depends, however, on the application context. 
If the goal is to analyze a single dataset and no further estimation is required, perhaps direct optimization via LSQ may be more practical, as it avoids the initial training phase of NBE. 
On the other hand, when estimation needs to be performed repeatedly, for example across multiple datasets, or large-scale simulation studies, the upfront training cost of NBE becomes negligible compared to the efficiency and accuracy gains at the inference stage. 
In such settings, NBE provides both faster and, as our results demonstrate, more accurate estimates.
We refer the reader to Section 2 of the review paper \cite{zammit2025} which discusses this trade-off between training cost and inference efficiency.

We further demonstrated the utility of NBE on high-frequency cryptocurrency return data. 
The method effectively captured temporal changes in process parameters, responded sensitively to market events, and offered robust uncertainty quantification. 
In particular, we considered two complementary approaches for UQ: a nonparametric bootstrap procedure, which captures the sampling variability of the estimator, and a posterior-quantile approach, which directly approximates posterior quantiles and allows construction of credible intervals in a likelihood-free manner.

These findings highlight NBE’s capacity to adapt to the nonstationary nature of real-world financial data while delivering stable and interpretable estimates: we obtained parameter estimates for high-frequency 1-minute bitcoin data over 9 years in matter of seconds.

Nonetheless, certain limitations remain. 
Accurately estimating parameters associated with deep subordination, such as those in the level-2 DVG model, proved more difficult in low-signal regimes. 
Future work may focus on improving estimation for these challenging regimes, possibly by incorporating structured priors or attention mechanisms to enhance model sensitivity. 
In addition, future extensions could include more refined posterior approximations or alternative bootstrap strategies to enhance uncertainty assessment.
Future directions may also include generalizing the method to multivariate L\'evy processes, or exploring extrapolation beyond the prior distribution of the parameters.

In conclusion, NBE demonstrates a strong ability to capture parameter dynamics in high-frequency cryptocurrency data. 
Its ability to deliver reliable inference with minimal computational cost makes it a powerful tool for modeling and analyzing complex financial systems.

\section*{Acknowledgements}

The authors thank the reviewers, associate editor and editor for constructive suggestions which improved the quality of the manuscript. 
This research was supported by National Science Foundation grant DMS-2310487.

\section{Appendix}

\subsection{Proof of Theorem \ref{th:risk.consistency}} \label{sec:proofTheorem2}
We begin with the basic decomposition
\[
R(\hat \theta_{m,N}^\star) - R^\star
= \big(R(\hat\theta_m^\star) - R^\star\big)
+ \big(R(\hat \theta_{m,N}^\star) - R(\hat\theta_m^\star)\big),
\]
where $R^\star=R(\hat\theta^\star)$ is the true Bayes risk, $\hat\theta_m^\star$ denotes the population minimizer over the class $\mathcal F_m$, and $\hat\theta_{m,N}^\star$ its empirical counterpart based on $N$ samples.

Consider the first difference. 
By (A2) the Bayes estimator $\hat\theta^\star(\cdot)$ is permutation-invariant. The universality property of DeepSets (A3) ensures that for every $\varepsilon>0$ there exists $m$ and $\tilde\theta_m\in\mathcal F_m$ with 
\[
\sup_x \|\tilde \theta_m(x) - \hat\theta^\star(x)\| \le \varepsilon.
\]
Since $L$ is Lipschitz (A1), this implies for all $(\theta,x)$ that

\[
\big|L(\theta,\tilde \theta_m(x)) - L(\theta,\hat\theta^\star(x))\big|
\le C_L \|\tilde \theta_m(x) - \hat\theta^\star(x)\| \le C_L \varepsilon.
\]

Therefore,
\begin{equation} \nonumber
\begin{split}
\left| R(\tilde \theta_m(x))) - R(\hat\theta^\star(x))   \right| & =  \big| \int_{\Theta}\int_{\mathcal S} \left[ L(\theta,\tilde \theta_m(x)) - L(\theta,\hat\theta^\star(x)) \right] \,f(x\mid\theta)\,dx\,d\Omega(\theta). \big| \\
 & \leq  \int_{\Theta}\int_{\mathcal S} \left| L(\theta,\tilde \theta_m(x)) - L(\theta,\hat\theta^\star(x)) \right| \,f(x\mid\theta)\,dx\,d\Omega(\theta). \\
 & \leq C_L \varepsilon \int_{\Theta}\int_{\mathcal S}  \,f(x\mid\theta)\,dx\,d\Omega(\theta).\\
 & = C_L \varepsilon.
\end{split}
\end{equation}

\noindent Since $\hat\theta_m^\star$ is the minimizer of $R$ over $\mathcal F_m$,
\[
R(\hat\theta_m^\star) \le R(\tilde \theta_m) \le R(\hat\theta^\star) + C_L\varepsilon.
\]
Thus $0 \le R(\hat\theta_m^\star) - R^\star \le C_L\varepsilon$, and since $\varepsilon>0$ was arbitrary, we conclude
\[
\lim_{m\to\infty} \big(R(\hat\theta_m^\star) - R^\star\big) = 0.
\]

For the second difference, fix $m$ and compare the empirical and population minimizers. 
Write
\[
R(\hat\theta_{m,N}^\star) - R(\hat\theta_m^\star)
= \big(R(\hat\theta_{m,N}^\star) - R_N(\hat\theta_{m,N}^\star)\big)
+ \big(R_N(\hat\theta_{m,N}^\star) - R_N(\hat\theta_m^\star)\big)
+ \big(R_N(\hat\theta_m^\star) - R(\hat\theta_m^\star)\big).
\]
The middle term is nonpositive by definition of $\hat\theta_{m,N}^\star$ as an empirical risk minimizer. 
The rest of the terms can be bounded by $\sup_{\hat\theta\in\mathcal F_m} |R_N(\hat\theta) - R(\hat\theta)|$.
Moreover, by the uniform law of large numbers (A4), this supremum converges in probability to zero as $N\to\infty$.
Hence
\[
0 \le R(\hat\theta_{m,N}^\star) - R(\hat\theta_m^\star) \le 2 \sup_{\hat\theta\in\mathcal F_m} | R_N(\hat\theta) - R(\hat\theta)| \xrightarrow{P} 0 ,
\]
so the error vanishes as $N\to\infty$ for fixed $m$.

Finally, for any $\delta>0$, choose $m$ large enough that $R(\hat\theta_m^\star) - R^\star \le \delta$ by the approximation argument above. 
For this fixed $m$, choose $N$ large enough so that $R(\hat\theta_{m,N}^\star) - R(\hat\theta_m^\star) \le \delta$ with high probability. 
Combining the two bounds yields
\[
R(\hat\theta_{m,N}^\star) - R^\star \le 2\delta,
\]
and since $\delta>0$ was arbitrary, we conclude that
\[
R(\hat\theta_{m,N}^\star) \xrightarrow{P} R^\star,
\]
for $(m,N)\to \infty$.

\subsection{Simulation Studies} \label{sec:appSimulations}

We conducted a series of simulation studies to systematically evaluate the performance of the neural Bayes estimator under a range of experimental configurations. 

These experiments were designed to help inform key modeling choices including: the number of prior samples $K$, the number of datasets per sample $J$, the input dimension $n_t$, and the neural network architecture—specifically the aggregation function, activation function, and loss function. 
The final values were selected based on preliminary trials and general performance stability, as discussed below.

To train the estimator across a broad range of plausible values, we selected reasonably uninformative prior distributions for the parameters of each process, similar as the priors used in Section ~\ref{sec:simDVG}. 
These were chosen to reflect realistic yet flexible ranges for each model's parameter space.
\begin{itemize}
    \item \textbf{Compound Poisson process:}
    $\lambda \sim U(0.1, 1.3), \mu \sim U(-0.6, 0.6),  \sigma^2 \sim U(10^{-3},0.3)$

    \item \textbf{Merton jump-diffusion model:} $\mu \sim U(-0.8, 0.8), \sigma^2 \sim U(10^{-3}, 1)$, $\lambda \sim U(0.1, 1.5)$, $\mu_J \sim U(-1.5,1.5), \sigma_J^2 \sim U(0.1,1.7)$

    \item \textbf{Variance gamma process:} $\gamma \sim U(-1.5,1.5),  \sigma^2 \sim U(10^{-4}, 2),  \alpha \sim U(0.1,3)$

    \item \textbf{Deep variance gamma (level 2):} $\sigma^2 \sim U(10^{-6}, 3), \alpha_1, \alpha_2 \sim U(10^{-6},25)$
\end{itemize}

For each model, parameters were independently sampled from the prior distributions. 
Simulated datasets were generated using these parameter samples, and each sample was used to produce $J$ independent and identically distributed datasets of length $n_t$, forming the training and validation datasets.

\subsubsection{Robustness to Priors}

To assess the sensitivity of our estimates to the choice of prior, we conducted additional simulation experiments in which the prior distributions of each parameter was enlarged. 
As expected, the RMSE for the estimates tends to increase with wider priors, since the estimator now has bigger (in magnitude) plausible parameter values.
To account for this effect and better evaluate consistency across prior specifications, we introduce two complementary metrics:

\begin{itemize}
    \item \textbf{Normalized root-mean-square error (NRMSE):} 
    \[
    \mathrm{NRMSE}(\hat{\theta}) = \frac{\sqrt{\frac{1}{n}\sum_{i=1}^{n} (\hat{\theta}_i - \theta_i)^2}}{\theta_\mathrm{max} - \theta_\mathrm{min}},
    \]
    where $\theta_\mathrm{max}$ and $\theta_\mathrm{min}$ denote the upper and lower bounds of the prior for the parameter $\theta$. 
    Normalizing by the prior length allows us to assess whether estimation quality remains proportional to the size of the parameter space.

    \item \textbf{Mean absolute percentage error (MAPE):} 
    \[
    \mathrm{MAPE}(\hat{\theta}) = \frac{1}{n} \sum_{i=1}^{n} \frac{|\hat{\theta}_i - \theta_i|}{|\theta_i|},
    \]
    which provides a measure of average relative error for each parameter.
\end{itemize}

The importance of these metrics is that we argue the following. 
If the normalized RMSE and MAPE remain approximately stable under prior expansion, the method is robust and produces consistent estimates regardless of the specific prior choice. 

For the robustness analysis, we considered three increasingly wide prior intervals for the parameters of the compound Poisson process. In the first specification (interval 1), we set
$\lambda \sim U(0.1, 1.3), \mu \sim U(-0.6, 0.6), \sigma^2 \sim U(10^{-3},0.3)$.
To assess sensitivity, we then expanded the support of each parameter by a factor of five, yielding the second specification (interval 2),
$\lambda \sim U(0.1, 6.5), \mu \sim U(-3, 3),  \sigma^2 \sim U(10^{-3},1.5)$, 
and finally by a factor of ten for the third specification (interval 3), $\lambda \sim U(0.1, 13), \mu \sim U(-6, 6),   \sigma^2 \sim U(10^{-3},3)$.

Similarly, we conducted the experiment with the DVG process of level 2.
In the first specification (interval 1), we use the same bounds as in the applications section,
$ \sigma^2 \sim U(10^{-6}, 3), \alpha_1, \alpha_2 \sim U(10^{-6},25).$
Then, we expanded the support of each parameter by a factor of five, yielding the second specification (interval 2),
$\sigma^2 \sim U(10^{-6}, 6), \alpha_1, \alpha_2 \sim U(10^{-6},50)$.

As reported in Tables ~\ref{tab:combined_results_poisson} and ~\ref{tab:combined_results_dvg}, while the absolute RMSE increases with wider priors, the NRMSE and MAPE for NBE remain relatively constant, suggesting that the neural Bayes estimator maintains reliable performance even under prior expansion. 
This provides empirical evidence of robustness to prior specification.
Moreover, the training times across the different settings were essentially comparable and thus did not represent a meaningful change in computational cost.

\begin{table}[t]
    \centering
    \scriptsize
    \begin{tabular}{|c|c|c|c|c|c|c|}
      \hline
       \textbf{Parameter/metric} & \multicolumn{2}{c|}{\textbf{Interval 1}} & \multicolumn{2}{c|}{\textbf{Interval 2}} & \multicolumn{2}{c|}{\textbf{Interval 3}} \\
      \hline
      
      \hline
      \textbf{$\lambda$} & \textbf{NBE} & \textbf{LSQ} & \textbf{NBE} & \textbf{LSQ} & \textbf{NBE} & \textbf{LSQ} \\
      \hline
      RMSE           & \textbf{0.051} & 0.058 & \textbf{0.33} & 1.38 & \textbf{1.11} & 3.96 \\
      Bias           & \textbf{0.0005} & -0.004 & \textbf{-0.0042} & 0.52 & \textbf{-0.012} & 1.54 \\
      NRMSE          & \textbf{0.051} & 0.058 & \textbf{0.053} & 0.22 & \textbf{0.087} & 0.31 \\
      MAPE           & \textbf{0.052} & 0.057 & \textbf{0.063} & 0.26 & \textbf{0.1} & 0.66 \\
      \hline
      
      \hline
      \textbf{$\mu$} & \textbf{NBE} & \textbf{LSQ} & \textbf{NBE} & \textbf{LSQ} & \textbf{NBE} & \textbf{LSQ} \\
      \hline
      RMSE           & \textbf{0.028} & 0.23 & \textbf{0.13} & 2.47 & \textbf{0.31} & 4.71 \\
      Bias           & \textbf{-0.005} & -0.02 & \textbf{-0.045} & -0.05 & -0.06 & \textbf{-0.05} \\
      NRMSE          & \textbf{0.023} & 0.19 & \textbf{0.021} & 0.41 & \textbf{0.026} & 0.39 \\
      MAPE           & \textbf{0.22} & 0.46 & \textbf{0.16} & 1.22 & \textbf{0.18} & 3.44 \\
    
      \hline
      
      \hline
      \textbf{$\sigma^2$} & \textbf{NBE} & \textbf{LSQ} & \textbf{NBE} & \textbf{LSQ} & \textbf{NBE} & \textbf{LSQ} \\
      \hline
      RMSE           & \textbf{0.022} & 0.13 & \textbf{0.20} & 0.37 & \textbf{0.62} & 1.23 \\
      Bias           & \textbf{7.6e-5} & 0.12 & \textbf{0.038} & 0.0019 & \textbf{-0.06} & -0.56 \\
      NRMSE          & \textbf{0.092} & 0.55 & \textbf{0.14} & 0.26 & \textbf{0.21} & 0.42 \\
      MAPE           & \textbf{0.11} & 0.97 & \textbf{0.39} & 0.71 & \textbf{0.72} & 0.87 \\
      \hline
    \end{tabular}
    \caption{Comparison of RMSE, Bias, NRMSE, and MAE (averaged over 1000 simulated datasets from the prior) for the parameters of the compound Poisson process across three intervals . The results are averaged and obtained using $nt=500$, $K=5000$, and $J=10$.
    }
    \label{tab:combined_results_poisson}
\end{table}

\begin{table}[t]
    \centering
    \scriptsize
    \begin{tabular}{|c|c|c|c|c|}
      \hline
       \textbf{Parameter/metric} & \multicolumn{2}{c|}{\textbf{Interval 1}} & \multicolumn{2}{c|}{\textbf{Interval 2}}  \\
      \hline
      
      \hline
      \textbf{$\sigma^2$} & \textbf{NBE} & \textbf{LSQ} & \textbf{NBE} & \textbf{LSQ} \\
      \hline
      RMSE           & \textbf{0.051} & 0.08 & \textbf{0.20} & 1.25  \\
      Bias           & \textbf{0.0024} & 0.0031 & \textbf{-0.011} & 0.008  \\
      NRMSE          & \textbf{0.017} & 0.02 & \textbf{0.014} & 0.08  \\
      MAPE           & \textbf{0.025} & 0.02 & \textbf{0.023} & 0.073  \\
      \hline
      
      \hline
      \textbf{$\alpha_1$} & \textbf{NBE} & \textbf{LSQ} & \textbf{NBE} & \textbf{LSQ}  \\
      \hline
      RMSE           & \textbf{6.72} & 9.63 & \textbf{35.34} & 53.8 \\
      Bias           & \textbf{-2.17} & -0.66 & \textbf{-11.57} & 6.69  \\
      NRMSE          & \textbf{0.27} & 0.39 & \textbf{0.28} & 0.43 \\
      MAPE           & \textbf{0.75} & 1.78 & \textbf{1.00} & 2.35  \\
    
      \hline
      
      \hline
      \textbf{$\alpha_2$} & \textbf{NBE} & \textbf{LSQ} & \textbf{NBE} & \textbf{LSQ} \\
      \hline
      RMSE           & \textbf{6.72} & 10.4 & \textbf{36.36} & 53.12  \\
      Bias           & \textbf{-2.54} & 2.3 & \textbf{-12.68} & 9.8  \\
      NRMSE          & \textbf{0.27} & 0.42 & \textbf{0.29} & 0.42  \\
      MAPE           & \textbf{0.66} & 1.00 & \textbf{0.99} & 2.06  \\
      \hline
    \end{tabular}
    \caption{Comparison of RMSE, Bias, NRMSE, and MAE (averaged over 1000 simulated datasets from the prior) for the parameters of a DVG process of level 2 across two intervals. The results are averaged and obtained using $nt=500$, $K=5000$, and $J=10$.
    }
    \label{tab:combined_results_dvg}
\end{table}

\subsubsection{Experiments on $n_t$}

We began our investigation by fixing $K = 5{,}000$ and $J = 10$, and exploring the impact of $n_t$—the number of observations per dataset—on both training time and estimation accuracy. 
Specifically, we evaluated how varying $n_t$ affects the root mean squared error (RMSE) of the estimated parameters and the overall training time of the neural network.

To test robustness, we repeated this analysis across the three  L\'evy process models discussed in Section ~\ref{sec:simpleModels}. 
For each model and each value of $n_t$, the estimator was trained on $K \cdot J$ datasets and then evaluated on an independent test set consisting of $1,000$ simulated datasets.

\begin{figure}[t]
 \centering
    \includegraphics[width=\textwidth]{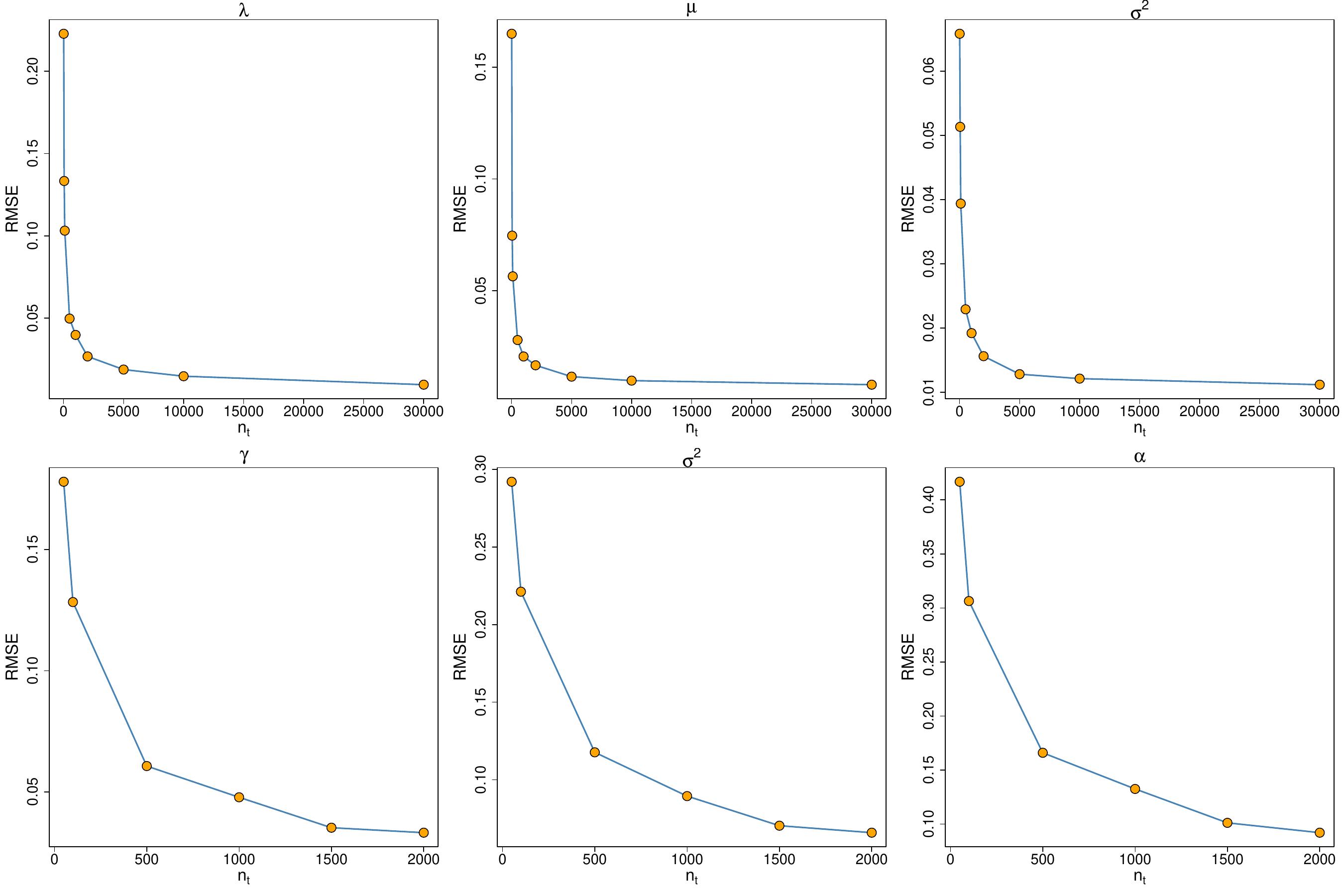}
  \caption{Root Mean Squared Error (RMSE) of parameter estimates for the Compound Poisson (top row) and Variance Gamma (bottom row) processes as a function of input length $n_t$.
  Each sub-plot corresponds to a different model parameter, as labeled.}
    \label{fig:rmseVsNt}
\end{figure}

\begin{figure}[t]
 \centering
    \includegraphics[width=\textwidth]{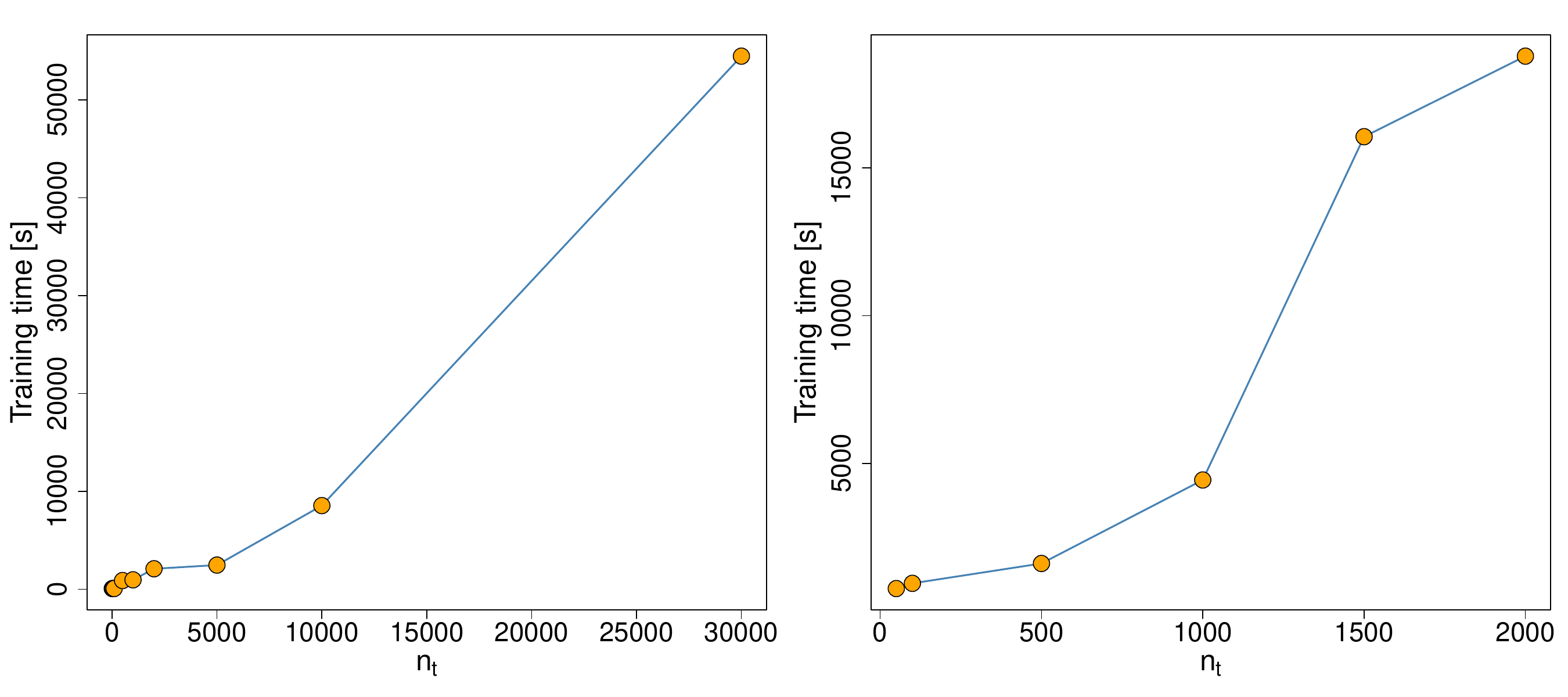}
  \caption{Time in seconds to train the NBE for the Compound Poisson (left) and Variance Gamma (right) processes as a function of input length $n_t$.}
    \label{fig:ttimeVsNt}
\end{figure}

To visualize the results, we plotted RMSE against $n_t$ for each parameter in each model in Figure~\ref{fig:rmseVsNt}. 
These plots help identify the point at which increasing the input length yields little or no additional improvement in accuracy. 
Across all models, we observe that RMSE steadily improves as $n_t$ increases, with diminishing returns beyond approximately $3{,}000$–$5{,}000$ observations. 
The figure also presents results for two different models—the compound Poisson and variance gamma processes—demonstrating similar trends. 
In fact, for the variance gamma process (bottom row), the point of diminishing returns appears even earlier, around $1{,}500$ observations.
Figure \ref{fig:ttimeVsNt} also shows how the training time increases as a function of $n_t$.

Given these findings, we selected $n_t = 1{,}440$ in our final experiments. 
This value corresponds to the number of one-minute log returns in a single trading day (24 hours), making it a natural and comparable choice for our cryptocurrency data application. 
While it lies slightly below the saturation threshold, it strikes a practical balance between empirical performance and application-specific relevance.

\subsubsection{Experiments on $K$}

We now investigate the impact of varying $K$, the number of parameter vectors sampled from the prior distribution. 

While we initially performed these studies across multiple L\'evy processes, we observed consistent trends in results across models. 
To reduce computational burden, we focused further tuning exclusively on the compound Poisson process, which served as a representative case. 
The final configurations reported in the main paper were then applied across all models.

\begin{figure}[t]
 \centering
    \includegraphics[width=\textwidth]{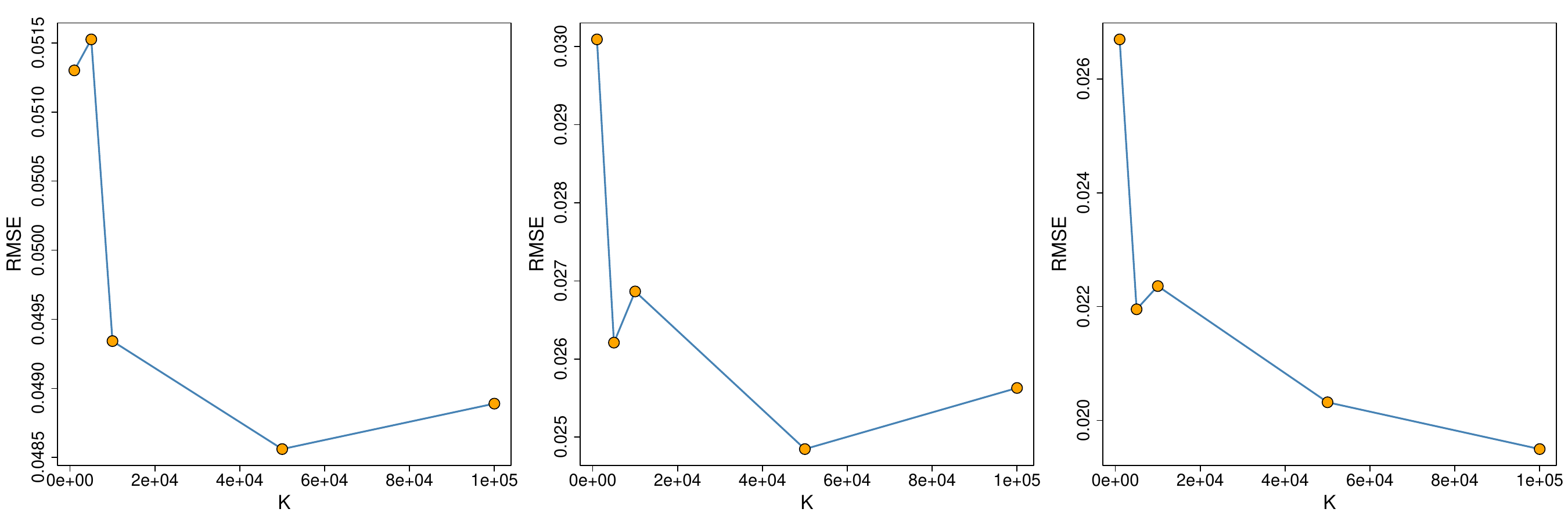}
  \caption{RMSE of parameter estimates as a function of $K$, the number of parameter vectors drawn from the prior. Performance improves with larger $K$, but exhibits diminishing returns beyond $10{,}000$.}
    \label{fig:rmse_vs_k}
\end{figure}

\begin{figure}[t]
 \centering
    \includegraphics[width=.7\textwidth]{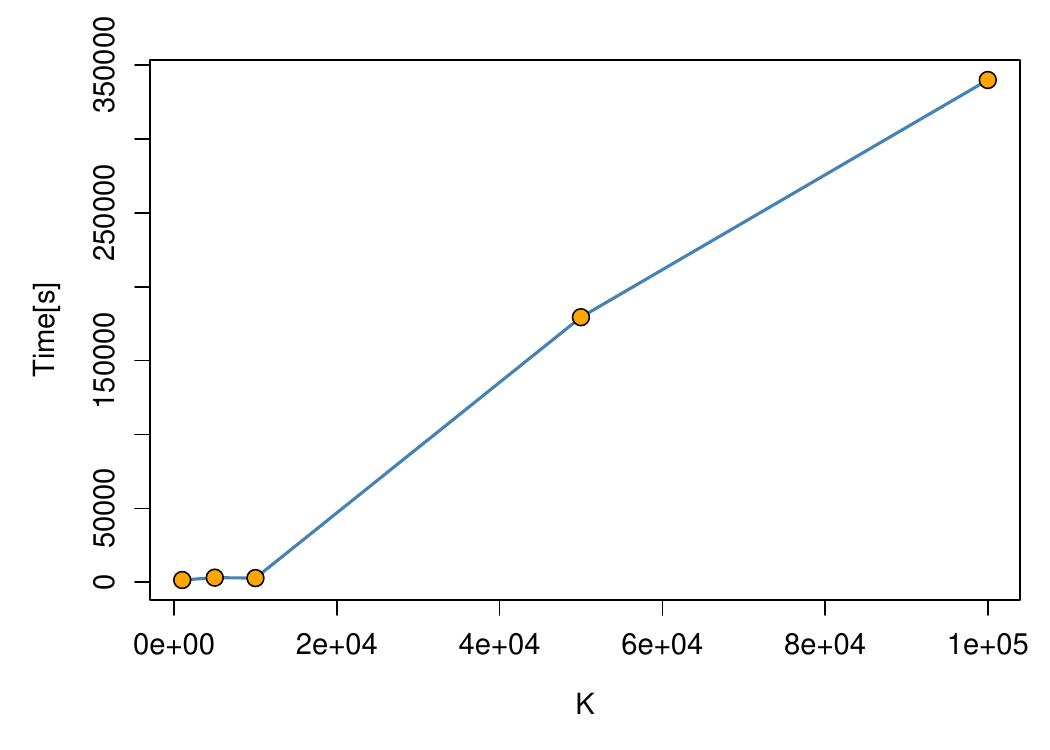}
  \caption{Training time (in seconds) as a function of $K$. Runtime increases rapidly, motivating a moderate choice of $K$ to maintain efficiency.}
    \label{fig:time_vs_k}
\end{figure}

Figure~\ref{fig:rmse_vs_k} shows how the RMSE of parameter estimates changes with increasing values of $K$. 
We observe that estimation error steadily decreases with larger $K$, but begins to plateau around $K = 50{,}000$. 
Beyond this point, the marginal gains in accuracy are minimal, sometimes the error even seems to increase.

However, as shown in Figure~\ref{fig:time_vs_k}, the training time increases rapidly with $K$. 
Based on these findings, we selected $K = 10{,}000$ for our final model. 
This value offers a strong trade-off between training efficiency and estimation accuracy.

We also emphasize again that the choice of $K$ is ultimately model-dependent. 
In particular, the required $K$ tends to increase with the number of parameters being estimated and the overall complexity of the model. 
The simulation experiments reported here serve as benchmarks for specific L\'evy processes, and may not directly generalize to all settings. 
For further guidance on selecting $K$, and additional simulation studies, we refer the reader to \citet{sainsbury2023nbe}.

As for $J$, we kept it fixed at $J = 10$ throughout all experiments. 
This choice is consistent with prior work \citep{sainsbury2023nbe}, which suggests that "$J$ can  be kept small (on the order of $10^0-10^1$) since data are simulated  for every sampled parameter vector".

\subsubsection{Experiments on Neural Network Configuration}

In parallel with the previous experiments, we investigated the impact of key architectural choices in the neural network design. 

We focused on the effect of two specific hyperparameters: the aggregation function and activation function.
While we fixed a standard architecture of three hidden layers with 32 neurons each for both the summary and inference networks.
This tuning phase was essential for understanding the estimator’s stability and sensitivity to design decisions.

For each configuration, we trained the model using $K = 10{,}000$ parameter samples and $J = 10$ datasets per parameter, as described earlier. 
After training, the network was evaluated on an independent test set of $1,000$ parameter samples drawn from the same prior. 
 
All experiments in this section were conducted using the level-2 DVG process.

\begin{table}[h]
    \centering
    \begin{tabular}{|c|c|c|c|}
      \hline
         \textbf{Agg. Function} &  $\sigma^2$ & 
      $\alpha_1$ & $\alpha_2$ \\
      \hline 
      Mean &  \textbf{0.074} & 6.61 & \textbf{6.65} \\
        Sum &  0.09 & \textbf{6.59} & 6.67 \\
        Max &  0.23 & 7.08 & 6.96 \\
        Min &  0.24 & 7.07 & 7.02 \\
      Product  & 0.88 & 7.17 & 7.16 \\
      \hline
\hline
          \textbf{Act. Function} &  $\sigma^2$ & 
      $\alpha_1$ & $\alpha_2$ \\
      \hline 
       RELU &  0.074 & 6.61 & 6.66 \\
      TANH &  0.10 & 6.61 & 6.68 \\
        WRELU &  \textbf{0.074} & \textbf{6.60} & \textbf{6.59} \\
      \hline

    \end{tabular}
    \caption{Root mean squared error (RMSE) for each parameter under different aggregation and activation functions. When testing one component, the others were held fixed.}
    \label{tab:avgscores}
\end{table}

Table~\ref{tab:avgscores} summarizes the RMSE results for each configuration across the three estimated parameters: $\sigma^2$, $\alpha_1$, and $\alpha_2$. 
When testing one component (e.g., activation), the other components were held fixed. 
In particular:
\begin{itemize}
    \item When comparing aggregation functions, we fixed the loss function to mean absolute error (MAE) and the activation function to Leaky ReLU.
    \item When testing activation functions, we fixed the aggregation to mean and the loss function to MAE.
\end{itemize}
In summary, based on the results of the simulation studies described in this appendix, we selected the following configuration for all experiments in the main text. 
We fixed the number of parameter draws to $K = 10{,}000$ and the number of replicates per parameter to $J = 10$, resulting in a training set of $100{,}000$ datasets. 
The input dimension $n_t$ was set to $1{,}440$, corresponding to one-minute returns over a full day.
For the neural network architecture, we used three hidden layers with 32 neurons per layer in both the summary and inference networks. 
Finally, we adopted the mean aggregation function, Leaky ReLU as the activation function, and mean squared logarithmic error (MSLE) as the loss function. 
This configuration demonstrated strong performance across all models considered.

\bibliography{bibliography-file}

\end{document}